\documentclass{article}

\usepackage[english]{babel}
\usepackage{amssymb}
\usepackage{amsfonts}
\usepackage{amsthm}
\usepackage{natbib}
\usepackage{tcolorbox}
\usepackage{multirow}
\usepackage{adjustbox}
\usepackage[affil-it]{authblk}
\usepackage[titletoc,title]{appendix}
\usepackage{amsmath}
\usepackage{relsize}
\usepackage{mathtools}
\usepackage{bbm}
\usepackage{tikz}
\usepackage{tcolorbox}

\DeclarePairedDelimiter\floor{\lfloor}{\rfloor}

\DeclareMathOperator*{\argmax}{arg\,max}
\DeclareMathOperator*{\argmin}{arg\,min}

\theoremstyle{definition}
\newtheorem{definition}{Definition}[section]
\newtheorem{proposition}{Proposition}[section]

\usepackage[letterpaper,top=2cm,bottom=2cm,left=3cm,right=3cm,marginparwidth=1.75cm]{geometry}

\usepackage{amsmath}
\usepackage{graphicx}
\usepackage[colorlinks=true, allcolors=blue]{hyperref}

\newcommand{\RN}[1]{ \textup{\uppercase\expandafter{\romannumeral#1}} }
 
\title{A Survey on Statistical Theory of Deep Learning: Approximation, Training Dynamics, and Generative Models
\footnote{This draft is a third revised version of our previous arXiv upload (July 4th). This article will appear in ``The Annual Review of Statistics and Its Application''.}}
\author{Namjoon Suh and Guang Cheng}

\affil{Department of Statistics and Data Science, UCLA}

\begin{document}
\maketitle

\begin{abstract}
In this article, we review the literature on statistical theories of neural networks from three perspectives: approximation, training dynamics and generative models.
In the first part, results on excess risks for neural networks are reviewed in the nonparametric framework of regression (and classification in Appendix~{\color{blue}B}). 
These results rely on explicit constructions of neural networks, leading to fast convergence rates of excess risks. Nonetheless, their underlying analysis only applies to the global minimizer in the highly non-convex landscape of deep neural networks. This motivates us to review the training dynamics of neural networks in the second part. Specifically, we review papers that attempt to answer ``how the neural network trained via gradient-based methods finds the solution that can generalize well on unseen data.'' In particular, two well-known paradigms are reviewed: the Neural Tangent Kernel (NTK) paradigm, and Mean-Field (MF) paradigm.
Last but not least, we review the most recent theoretical advancements in generative models including Generative Adversarial Networks (GANs), diffusion models, and in-context learning (ICL) in the Large Language Models (LLMs) from two perpsectives reviewed previously, i.e., approximation and training dynamics.
\end{abstract}

\maketitle

\section{Introduction}
In recent years, the field of deep learning~\citep{goodfellow2016deep} has experienced a substantial evolution. 
Its impact has transcended traditional boundaries, leading to significant advancements in sectors such as healthcare~\citep{esteva2019guide}, finance~\citep{heaton2017deep}, autonomous systems~\citep{grigorescu2020survey}, and natural language processing~\citep{otter2020survey}. 
Neural networks, the mathematical abstractions of our brain, lie at the core of this progression.
Nevertheless, amid the ongoing renaissance of artificial intelligence, neural networks have acquired an almost mythical status, spreading the misconception that they are more art than science. 
It is important to dispel this notion. 
While the applications of neural networks may evoke awe, they are firmly rooted in mathematical principles.
In this context, the importance of deep learning theory becomes evident. 
Several key points underscore its significance.

\subsection{Why theory is important?} \label{theory_import}
In this subsection, we aim to emphasize the importance of understanding deep learning within mathematical and statistical frameworks. 
Following are some key points to consider:

\begin{enumerate}
    \item Deep learning is a dynamic and rapidly evolving field, producing hundreds of thousands of publications online. Today's models are characterized by highly intricate network architectures comprising numerous complex sub-components (e.g., Transformer~\citep{vaswani2017attention}). Amidst this complexity, it becomes crucial to comprehend the fundamental principles underlying these models. To achieve this understanding, placing these models within a unified mathematical framework is essential. Such a framework serves as a valuable tool for distilling the core concepts from these intricate models, making it possible to extract and comprehend the key principles that drive their functionality.
    
    \item  Applying statistical frameworks to deep learning models allows meaningful comparisons with other statistical methods. For instance, widely used statistical estimators like wavelet or kernel methods can prompt questions about when and why deep neural networks might perform better. This analysis helps us understand when deep learning excels compared to traditional statistical approaches, benefiting both theory and practice. 

    \item Hyperparameters, such as learning rate, weight initializations, network architecture choices, activation functions, etc, significantly influence the quality of the estimated model. Understanding the proper ranges for these hyperparameters is essential, not only for theorists but also for practitioners. For instance, in the era of big data where there are millions of samples in one dataset, the theoretical wisdom gives us the depth of the network should scale logarithmically in sample size for the good estimation of compositional functions 
    (e.g., See~\cite{schmidt2020nonparametric}).
\end{enumerate}
In this review, we provide an overview of papers that delve into these concepts within mathematical settings, offering readers specific insights into the topics discussed above.
Here, we try to avoid too many technicalities and make the introductions accessible to as many statisticians as possible. 
Some overly technical components are deferred to the Appendix. 

\subsection{Roadmap of the paper}
We classify the existing literature on statistical theories of neural networks into three categories. (Section~\ref{Approx_theory}, Section~\ref{training_dynamics}, Section~\ref{Generative_Models}, respectively.)

\begin{enumerate}
    \item \noindent \textit{\textbf{Approximation theory viewpoint.}}
    Recently, there has been a huge collection of works which bridge the approximation theory of neural network  models~\citep{yarotsky2017error,mhaskar1996neural,petersen2018optimal,schmidt2020nonparametric,montanelli2019new,blanchard2021shallow,hornik1989multilayer,hanin2019universal} with the tools in empirical processes~\citep{van2000empirical} to obtain the fast convergence rates of excess risks both in regression~\citep{schmidt2020nonparametric, hu2021regularization} and classification~\citep{hu2020optimal, kim2021fast} tasks under non-parametric settings.
    Approximation theory provides useful perspectives in measuring the fundamental complexities of neural networks for approximating functions in certain classes. 
    Specifically, it enables the explicit constructions of neural networks for the function approximations so that we know how the network width, depth, and number of active parameters should scale in terms of sample size, data dimension, and the function smoothness index to get good statistical convergence rates.     
    For simplicity, we mainly consider the works in which the fully connected neural networks are used as the function estimators.
    These works include~\cite{schmidt2020nonparametric,kim2021fast,shen2019deep,jiao2021deep,lu2021deep,imaizumi2019deep,imaizumi2022advantage,suzuki2018adaptivity,chen2019nonparametric,suzuki2021deep,suh2022approximation} under various problem settings. 
    Yet, these works assume that the global minimizers of loss functions are obtainable, and are mainly interested in the statistical properties of these minimizers without any optimization concerns.
    However, this is a strong assumption, given the non-convexity of loss functions arising from the non-linearities of activation functions in the hidden layers.
    

    \item \noindent \textit{\textbf{Training dynamics viewpoint. }}
    Understanding the landscape of non-convex loss functions for neural network models and its impact on their generalization capabilities represents a critical next step in the literature. However, the non-trivial non-convexity of this landscape poses significant challenges for the mathematical analysis of many intriguing phenomena observed in neural networks.
    For example, a seminal empirical finding,~\cite{zhang2021understanding}, reveals that neural networks in their experiments trained on a standard image classification training set (CIFAR 10), can fit the (noisy) training data perfectly, but at the same time showing respectable prediction performance. (See Figure 1(c) in~\cite{zhang2021understanding}.)
    This contradicts the classic statistical wisdom of bias-variance tradeoff, which states the overfitted models cannot generalize well. 
    The role of overparameterizations (e.g.,~\cite{bartlett_montanari_rakhlin_2021}) on the non-convex optimization landscape of neural networks has been intensively studied over the past years, and we review the relevant literatures under this context. 
    For instance,~\cite{jacot2018neural} revealed that the dynamics of highly overparameterized neural networks with large enough width, trained via gradient descent (GD) in $\ell_{2}$-loss, behave similarly to those of functions in reproducing kernel Hilbert spaces (RKHS), where the kernel is associated with a specific network architecture.
    Many subsequent works study the training dynamics and the generalization abilities of neural networks in the kernel regime under various settings~\citep{hu2021regularization,nitanda2020optimal}.
    However, due to technical constraints (as detailed in Subsection~\ref{NTK_sec}), networks in the kernel regime fail to explain the essential functionality of neural networks—\textit{feature learning}~\citep{zhong2016overview}.    
    Another important line of work focuses on understanding the learning dynamics of neural networks in the Mean-Field (MF) regime, where feature learning becomes more explainable. 
    Nonetheless, the analysis within the MF regime is challenging to generalize to deep networks and requires infinite widths.
    Finally, we conclude this section by presenting several approaches that go beyond or unify the two regimes.
    
    \item \noindent \textit{\textbf{Generative modeling.}} 
    In this section, we review the most recent theoretical advancements in generative models including \textit{Generative Adversarial Networks (GANs)}, \textit{Diffusion Models}, and \textit{In-context learning} in Large Language Models (LLMs).
    The works to be introduced are based on the philosophies of two paradigms (approximation \& training dynamics).
    Over the past decade, GANs~\citep{goodfellow2014generative} have stood out as a significant unsupervised learning approach, known for their ability to learn 
    the data distributions and efficiently sample the data from it.
    In this review, we will introduce papers that studied statistical properties of GANs~\citep{arora2017generalization, liang2021well, chen2020distribution, bai2018approximability, zhang2017discrimination, schreuder2021statistical}.
    Recently, another set of generative models, i.e., diffusion model, has shown superior performances in generating impressive qualities of synthetic data in various data modalities including image~\citep{song2020score, dhariwal2021diffusion}, tabular data~\citep{kim2022stasy,suh2023autodiff}, medical imaging~\citep{muller2022diffusion} etc., beating GAN based models by a large margin. However, given the model's complex nature and its recent introduction in the community, the theoretical reason why it works so well remains vague. Lastly, we will review an interesting phenomenon commonly observed in Large Language Models referred to as {In-context learning} (ICL). It refers to the ability of LLMs conditioned on prompt sequence consisting of examples from a task (input-output pairs) along with the new query input, the LLM can generate the corresponding output accurately. 
    Readers can refer to nice survey papers~\cite{gui2021review,yang2022diffusion} on the detailed descriptions of methodologies and the applications of GANs, and diffusion models in various domains.
    For an overview of ICL, see survey by~\cite{dong2022survey} which highlights some key findings and advancements in this direction.
\end{enumerate}
In relation to Subsection~\ref{theory_import}, the advantages of neural networks over classic statistical function estimators are primarily discussed in Sections~\ref{Approx_theory} and~\ref{training_dynamics} under various problem settings. 
In Section~\ref{training_dynamics}, we review the work of~\cite{yang2020feature}, which suggests appropriate parameter initialization scalings and learning rates for feature learning in large-scale (infinite width) neural networks.

\subsection{Existing surveys on deep learning theory}
To our knowledge, there are four existing survey papers~\citep{bartlett_montanari_rakhlin_2021, fan2021selective, belkin2021fit, he2020recent} on deep learning theory.
There is an overlap in certain subjects covered by each of the papers, but their main focuses are different from each other.
\cite{bartlett_montanari_rakhlin_2021} provided a comprehensive and technical survey on statistical understandings of deep neural networks. 
In particular, the authors focused on examining the significant influence of overparameterization in neural networks, which plays a key role in enabling gradient-based methods to discover interpolating solutions. 
\cite{fan2021selective} introduced the most commonly employed neural network architectures in practice such as Convolutional Neural Net (CNN), Recurrent Neural Networks (RNN), and training techniques such as batch normalization, and dropout.
A brief introduction to the approximation theory of neural networks is provided as well.
Similarly as~\cite{bartlett_montanari_rakhlin_2021},~\cite{belkin2021fit} reviewed the role of overparameterizations for implicit regularization and benign overfitting observed not only in neural network models but also in classic statistical models such as weighted nearest neighbor predictors. 
Most notably, they provided intuitions on the roles of the overparameterization of non-convex loss landscapes of neural networks through the lens of optimization.
\cite{he2020recent} provides a comprehensive overview of deep learning theory, including the 
ethics and security problems that arise in data science and their relationships with deep learning theory.
We recommend readers review all these papers to gain a comprehensive understanding of this emerging field.
Our paper offers a unique and comprehensive survey of the statistical results of neural networks, focusing on approximation theory and training dynamics, while also covering generative models within these two paradigms.

\section{Approximation theory-based statistical guarantees} \label{Approx_theory}
We outline fully connected networks, which are the main object of interest throughout this review. 
From a high level, deep neural networks can be viewed as a family
of nonlinear statistical models that can encode highly nontrivial representations of
data.
The specific network architecture $(L,\mathbf{p})$ consists of a positive integer $L$, called \textit{the number of hidden layers}, and \textit{a width vector} 
$\mathbf{p}:=(\mathbf{p}_{0},\dots,\mathbf{p}_{L+1})\in\mathbb{N}^{L+2}$, recording the number of nodes from input to output layers in the network.
A fully connected neural network, $\Tilde{f}$, is then any function of the form for the input feature $\mathbf{x}\in\mathcal{X}$:
\begin{align} \label{DNN}
    \Tilde{f}:\mathcal{X}\rightarrow{\mathbb{R}}, \quad
    \mathbf{x}\rightarrow{f(\mathbf{x})={W}_{L}\mathbf{\sigma}{W}_{L-1}\mathbf{\sigma}{W}_{L-2}\dots\mathbf{\sigma}{W}_{1}\mathbf{x}},
\end{align}
where $\mathbf{W}_{i}\in\mathbb{R}^{p_{i+1} \times p_{i}}$ is a weight matrix with $\mathbf{p}_{0}=d$, $\mathbf{p}_{L+1}=1$ and $\mathbf{\sigma}$ is the non-linear activation function.
Here, the activation function plays a \textit{key-role} in neural network allowing the non-linear representations of the given data $\mathbf{x}$. 
Popular examples include: ReLU($x$)=$\max(x,0)$, Sigmoid($x$)=$\frac{1}{1+e^{-x}}$.
We omit the bias terms added on the outputs of pre-activated hidden layers for simplicity.
But bias terms are needed for universal approximation if the input data is not appended with a constant entry. 
\\ \\
Under this setting, complexity of the networks is mainly measured through the three metrics: \textit{$(1)$ the maximum width, denoted as $\mathbf{p}_{\textbf{max}}:=\max_{i=0,\dots,L+1}\mathbf{p}_{i}$}, \textit{$(2)$ the depth, denoted as $L$}, and {\textit{$(3)$ the number of non-zero parameters, denoted as $\mathcal{N}$}}.
Letting $\|\mathbf{W}_{j}\|_{0}$ be the number of non-zero entries of $\mathbf{W}_{j}$ in the $j^{\text{th}}$ hidden layer, the final form of neural network we consider is given by:
\begin{align} \label{network_def}
    \mathcal{F}(L,\mathbf{p},\mathcal{N}) := 
    \bigg\{ \Tilde{f} \text{ of the form~\eqref{DNN}} :
    \sum_{j=1}^{L}\|\mathbf{W}_{j}\|_{0} \leq \mathcal{N}
    \bigg\}.
\end{align}
In the approximation theoretic literature, the capacity or expressive power of neural network is often characterized by the tuple $(L,\mathbf{p}_{\textbf{max}},\mathcal{N})$.
Let $\mathcal{G}$ be a function class where the target function $f_{\star}$ belongs. 
The main question frequently asked is: given the fixed approximation error, $\varepsilon_{\text{Apprx}}$, defined as follows: 
\begin{align}
    \varepsilon_{\text{Apprx}}:=\sup_{f_{\star}\in\mathcal{G}} \inf_{f \in \mathcal{F}(L,\mathbf{p},\mathcal{N})} 
    \left\| f-f_{\star} \right\|_{L_{p}},
\end{align}
how does the network architecture $(L,\mathbf{p}_{\textbf{max}},\mathcal{N})$ scale in terms of $\varepsilon_{\text{Apprx}}$? (henceforth, the subscript $\textit{Apprx}$ will be omitted in $\varepsilon_{\text{Apprx}}$.)
Note the~\textit{sup} is taken over the function class $\mathcal{G}$ and the~\textit{inf} is taken over the neural network class $\mathcal{F}$.
The distance between two functions is measured via $L_{P}$-norm.

\subsection{Expressive power of fully-connected networks} \label{Expressive_power}
In this subsection, we briefly review some important results in the approximation theory of neural networks.
For more comprehensive reviews on this topic, readers can refer to~\cite{devore2021neural}.
\\ \\
\noindent \textbf{{Approximating functions in $\mathcal{G}$}}:  
The specifications of function class $\mathcal{G}$ and $\mathcal{F}$ allow us to derive many interesting insights on the power of neural networks.
For instance, the celebrated \textit{``universal approximation theorem''} states that any continuous functions (i.e., $\mathcal{G}:=\{\text{Continuous Functions on $\mathbb{R}^{d}$}\}$) can be approximated by a shallow neural network (i.e., one-hidden layer) with sigmoid activation function (i.e., $\mathcal{F}:=\{\text{Shallow Neural Networks}\}$) at an arbitrary accuracy (\cite{cybenko1989approximation, hornik1989multilayer, hornik1990universal}).
However, achieving a good approximation may require an extremely large number of hidden nodes, which significantly increases the capacity of $\mathcal{F}$.
~\cite{barron1993universal,barron1994approximation} developed an approximation theory for function classes $\mathcal{G}$ with limited capacity, measured by the integrability of their Fourier transform.
Interestingly, the approximation result is not affected by the dimension of input data $d$, and this observation matches with the experimental results that deep learning is very effective in dealing with high-dimensional data.
\\ \\ 
Nonetheless, the capacity of $\mathcal{G}$ in~\cite{barron1994approximation} is rather limited. 
Another typical route of the analysis is to specifying the smoothness of function classes. 
Readers can roughly understand that smoothness refers to the highest order of derivatives that the functions can possess. 
Notably,~\cite{yarotsky2017error} demonstrated that deep ReLU networks~\eqref{DNN} cannot escape~\textit{the curse of dimensionality} when approximating functions in the unit ball in Sobolev space. 
They established that the order $\mathcal{N} = \mathcal{O}(\varepsilon^{-\frac{d}{r}})$ is sharp, with matching lower and upper bounds. 
\cite{petersen2018optimal} generalized the results to the class of \textit{piece-wise} smooth functions. 
Later,~\cite{schmidt2020nonparametric} developed a theory for any network architecture satisfying the set of conditions on $(L,\mathbf{p}_{\textbf{max}},\mathcal{N})$, deep ReLU nets can achieve good approximation rates for functions in H\"older class. (More details on the technical results of~\cite{schmidt2020nonparametric} are deferred in the Appendix~{\color{blue}A}.)
This should be contrasted to the~\cite{yarotsky2017error} where they proved the ``existence'' of a network with good approximation. 
Many researchers have been working on considering either more general (i.e., Besov space) or more specific (i.e., hierarchical compositional function) function classes $\mathcal{G}$ than H\"older class.
These considerations have facilitated numerous intriguing comparisons between classic statistical function estimators and deep neural networks in terms of their fundamental limits, specifying the second item in Subsection~\ref{theory_import}. 
\\ \\
\noindent \textbf{{The benefits of depth}}: It has been shown that the expressive power of deep neural networks grows with respect to the number of layers (i.e., $L$) by several studies.
\cite{delalleau2011shallow} showed there exist families of functions that can be represented much more efficiently with a deep network than with a shallow one (i.e. with substantially fewer hidden units).
In the asymptotic limit of depth,~\cite{pascanu2013number} showed deep ReLU networks can represent exponentially more piece-wise linear functions than their single-layer counterparts can do, given that both networks have the same number of nodes.
Later,~\cite{montufar2014number} proved that a similar result can be derived with the fixed number of hidden layers. 
\cite{poole2016exponential} showed deep neural networks can disentangle highly curved manifolds in an input space into flat manifolds in a hidden space, while shallow networks cannot. 
\cite{mhaskar2017and} demonstrated that deep ReLU networks can approximate compositional functions with significantly fewer parameters (i.e., $\mathcal{N}$) than shallow neural networks need in order to achieve the same level of approximation accuracy.
\\ \\
\noindent \textbf{{Bounded-width}}:
The effects of width on the expressive power of neural networks have recently been studied~\citep{lu2017expressive, hanin2019universal, kidger2020universal, park2020minimum, vardi2022width}.
~\cite{lu2017expressive} showed that the minimal width for universal approximation (denoted as $w_{\text{min}}$) using ReLU networks w.r.t. the $L_{1}$ norm of functions from $\mathbb{R}^{d}\rightarrow\mathbb{R}$ is $d+1\leq w_{\text{min}} \leq d+4$.
~\cite{kidger2020universal} extended the results to $L_{p}$-approximation of functions from $\mathbb{R}^{d}\rightarrow\mathbb{R}^{\text{out}}$, and obtained $w_{\text{min}}\leq d+d_{\text{out}}+1$.
~\cite{park2020minimum} further improved $w_{\text{min}} = \max\{d+1, d_{\text{out}}\}$.
Universal approximations of narrow networks with other activation functions had been studied in~\cite{park2020minimum, kidger2020universal, johnson2018deep}.
Note that the aforementioned works require the depth of networks to be exponential in input dimension with bounded-width, which are the dual-versions of universal approximation of bounded depth networks from~\cite{cybenko1989approximation,hornik1989multilayer,hornik1990universal}. 
Interestingly,~\cite{vardi2022width} provided an evidence that width of the networks can be less important than depth for the expressive power of neural nets. 
They showed that the price for making the width small is only a linear increase in the network depth, in sharp contrast to the results mentioned earlier on how making the depth small may require an exponential increase in the network depth.


\subsection{Statistical guarantees for regression tasks} \label{Approx_guarantees}
The natural question is: what are the interpretations or consequences of the results in approximation theory for deriving statistical guarantees of neural networks under \textbf{\textit{noisy observations}}?
In this subsection, we focus on reviewing several important results under regression tasks in this regard.
As preliminary, we introduce the settings frequently adopted in statistical learning theory. 
\\ \\
Let $\mathcal{X}$ and $\mathcal{Y}\subset\mathbb{R}$ be the measurable feature space and output space.
We denote $\rho$ as a joint probability measure on the product space $\mathcal{X}\times\mathcal{Y}$, and
let $\rho_{\mathcal{X}}$ be the marginal distribution of the feature space $\mathcal{X}$.
We assume that the noisy data set $\mathcal{D}:=\{(\mathbf{x}_{i},\mathbf{y}_{i})\}_{i=1}^{n}$ are generated from the non-parametric regression model: 
\begin{align} \label{non_para}
    \mathbf{y}_{i}=f_{\rho}(\mathbf{x}_{i})+\varepsilon_{i},\quad i=1,2,\dots,n,
\end{align}
where the noise $\varepsilon_{i}$ is assumed to be centered random variable and $\mathbb{E}(\varepsilon_{i}|\mathbf{x}_{i})=0$.
Our goal is to estimate the regression function $f_{\rho}(\mathbf{x})$ with the given noisy data set $\mathcal{D}$.
Here, it is easy to see regression function $f_{\rho}:=\mathbb{E}(\mathbf{y}|\mathbf{x})$ is a minimizer of the  population risk $\mathcal{E}(f)$ under $\ell_{2}$-loss defined as: 
\begin{align*}
    \mathcal{E}(f):=\mathbb{E}_{(\mathbf{x},\mathbf{y})\sim\rho}\bigg[ \big(\mathbf{y}-f(\mathbf{x})\big)^2 \bigg].
\end{align*}
\noindent
However, since the joint distribution $\rho$ is unknown, we cannot find $f_{\rho}$ directly.
Instead, we solve the following empirical risk minimization problem induced from the dataset $\mathcal{D}$: 
\begin{align} \label{Minimizer}
    \widehat{f}_{n} =\argmin_{f\in\mathcal{F}(L,\mathbf{p},\mathcal{N})} \mathcal{E}_{D}(f)
    :=\argmin_{f\in\mathcal{F}(L,\mathbf{p},\mathcal{N})}\bigg\{ \frac{1}{n}\sum_{i=1}^{n}\big(\mathbf{y}_{i}-f(\mathbf{x}_{i})\big)^{2} \bigg\}.
\end{align}
Note that the papers introduced in this subsection always assume the empirical risk minimizer $\widehat{f}_{n}$ is obtainable, ignoring the optimization process. 
The function estimator $f$ is structurally regularized by $\mathcal{N}$ in $(L,\mathbf{p},\mathcal{N})$, which will be specified in sequel. 
\\ \\
Under this setting, the excess risk is an important statistical object measuring the generalizability of the function estimator $\widehat{f}_{n}$ for unseen data in $\mathcal{X}$.
Mathematically, it can be shown that it is a difference between population risks of $f_{\rho}$ and $\widehat{f}_{n}$ (See Chapter $13$ in~\cite{wainwright2019high}), which is $\mathbb{E}_{\mathbf{X}\sim\rho_{\mathcal{X}}}\big[(\widehat{f}_{n}(\mathbf{X})-f_{\rho}(\mathbf{X}))^{2}\big]$.
The excess risk can be further decomposed as follows (Proposition $4.2$ in \cite{suh2022approximation}) :
\begin{align} \label{Decomposition}
    \mathbb{E}_{\mathbf{X}\sim\rho_{\mathcal{X}}}\big[(\widehat{f}_{n}(\mathbf{X})-f_{\rho}(\mathbf{X}))^{2}\big] 
    \leq \frac{\text{Complexity Measure of $\mathcal{F}$}}{n} + \text{Approx. Error}^{2}. 
\end{align}
In the context of excess risk bounds, it is important to note the trade-off between the \textbf{\textit{“Approximation error”}} and the combinatorial \textbf{\textit{“Complexity measure”}} of a neural network class $\mathcal{F}$. Specifically, as the network hypothesis space $\mathcal{F}$ becomes richer, the approximation results improve. 
However, increasing the hypothesis space $\mathcal{F}$ arbitrarily will eventually lead to an increase in the complexity measure of $\mathcal{F}$, as described in~\eqref{Decomposition}. 
Researchers (e.g., \cite{bartlett2019nearly,schmidt2020nonparametric}) have examined how various complexity measures, including VC-dimension, pseudo-dimension, and covering number, scale with respect to $(L,\mathbf{p}_{\text{max}}, \mathcal{N})$. 
Specifically, above papers proved all the three complexity measures increase linearly in $\mathcal{N}$.
For achieving good convergence rates of the excess risks from~\eqref{Decomposition}, it is crucial to properly specify the network architecture (i.e., the choices of $(L,\mathbf{p}_{\text{max}},\mathcal{N})$) that balances the tension between \textit{complexity of $\mathcal{F}$} and \textit{approximation error} in terms of sample size $n$, data dimension $d$, and function smoothness $r \geq 0$.
\\ \\
\noindent \textbf{{Deep sparse ReLU networks v.s. Linear estimators}}: 
Among the list of papers to be introduced shortly, the seminal work~\cite{schmidt2020nonparametric} paved the way for providing the statistical guarantees of deep ReLU networks in the sense of~\eqref{Decomposition}. 
In their work, they demonstrated that \textbf{\textit{sparsely}} connected deep ReLU networks~\eqref{network_def} significantly outperform traditional statistical estimators.
Specifically, if the unknown regression function \( f_{\rho} \) is a composition of functions that are individually estimable faster than $\mathcal{O}\big(n^{-\frac{2r}{2r+d}}\big)$, then a composition-based deep ReLU network is provably more effective than estimators that do not utilize compositions, such as wavelet estimators.
For further discussions on the paper, interested readers can find the published discussion papers~\cite{kutyniok2020discussion,ghorbani2020discussion,shamir2020discussion,kohler2020discussion}.
\\ \\
The sparse network structure manifested in $\mathcal{N}$ in the paper had already been proven to be impressively effective in compressed learning literature ~\citep{iandola2016squeezenet,han2015deep,han2015learning}.
The sparsity of networks can be achieved via pruning technique~\citep{han2015learning}. One of them,~\cite{iandola2016squeezenet}, empirically showed the pruned convolutional neural networks with 50 times fewer parameters achieve the same accuracy level as that of the AlexNet~\citep{krizhevsky2012imagenet} in image classification tasks, and these results pave the way for the employments of neural networks in small devices such as smart phones or smart watches. 
\\ \\
\begin{figure}[!t]
  \centering
  \includegraphics[width=1\textwidth]{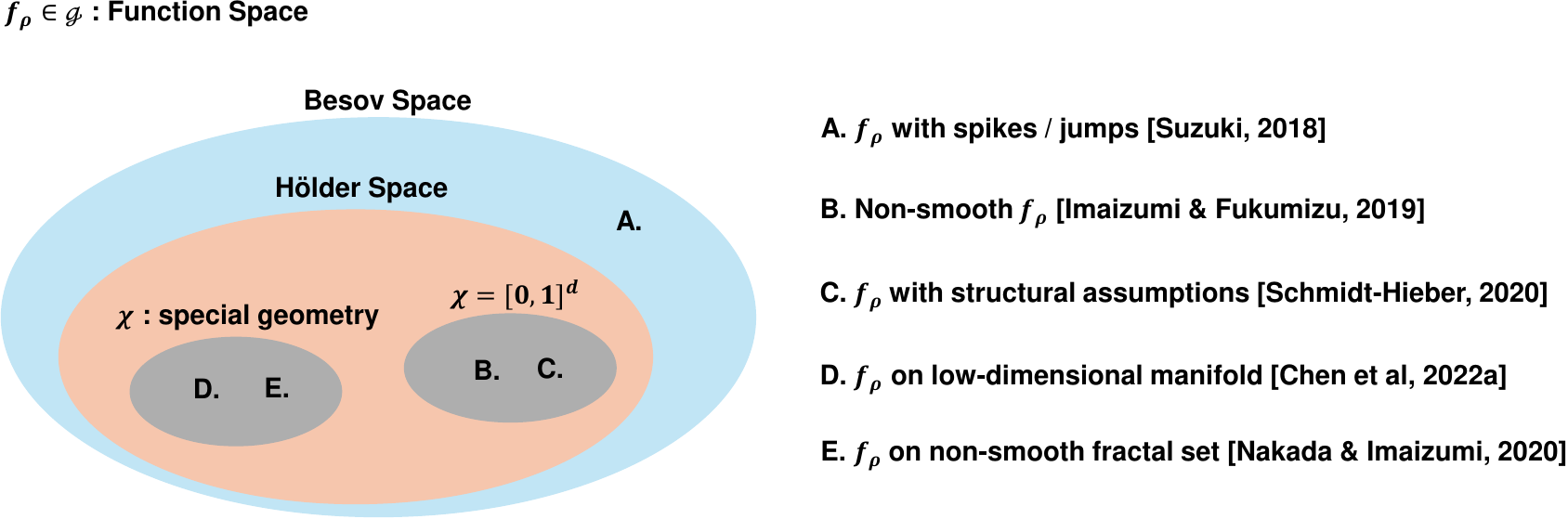}
  \caption{Compared to classical linear estimators (i.e., wavelet, kernel ridge regressors, etc), sparsely connected neural networks are more adaptive in estimating functions $f_{\rho}$ with special structures. 
  The figure illustrates the different settings of function classes $\mathcal{G}$ where neural networks exhibit superior adaptabilities over the classical estimators.}
  \label{Fig0}
\end{figure}
In statistical literature, after the first appearance of~\cite{schmidt2020nonparametric}, several works were followed subsequently analyzing sparsely connected networks.
\cite{imaizumi2019deep} derived the excess risk convergence rate of sparse ReLU neural networks estimating piece-wise smooth functions, showing that deep learning can outperform the classical linear estimators including kernel ridge regressors, Fourier estimators, splines, and Gaussian processes. 
(Here, we refer to the estimators as \textit{`linear'} when they are linearly dependent on the output $\mathbf{y}$.)
They pointed out the discrepancy between deep networks and linear estimators appears when the target function is \textit{non-smooth}. 
Another work,~\cite{suzuki2018adaptivity} showed the great adaptiveness of sparse ReLU networks (i.e.,\eqref{network_def}) for the functions in Besov space. 
The Besov space is a general function space including H\"older space.
Specifically, it also allows functions with spatially inhomogeneous smoothness with spikes and jumps. 
The author mentioned deep networks possess strong adaptiveness in capturing the spatial inhomogeneity of functions, whereas linear estimators are only able to capture the global properties of target functions and cannot capture the variability of local shapes.
Later,~\cite{hayakawa2020minimax} proved the linear estimators cannot distinguish the function class and its convex hull. 
This results in the suboptimality of linear estimators over a simple but non-convex function class, on which sparsely connected deep ReLU nets can attain nearly the minimax-optimal rate.
There also have been efforts~\citep{kohler2021rate, farrell2021deep} to study the statistical guarantees of \textit{\textbf{densely}} fully-connected networks without sparsity constraints. 
However, the rates of excess risk they obtained are sub-optimal.  
\\ \\
\noindent \textbf{{Avoiding curse of dimensionality}}: 
According to the classical result from \cite{donoho1998minimax}, for estimating functions in the H\"older class with smoothness $r\geq 0$, the un-improvable minimax convergence rate of excess risk is:
\begin{align} \label{excess_risk}
\inf_{\widehat{f}_{n}}\sup_{f_{\rho}\in\{\text{H\"older}\}}\mathbb{E}_{\mathbf{X}\sim\rho_{\mathcal{X}}}\big[(\widehat{f}_{n}(\mathbf{X})-f_{\rho}(\mathbf{X}))^{2}\big] = \mathcal{O}\bigg( n^{-2r/2r+d} \bigg). 
\end{align}
This rate can be problematic when the data dimension is much larger than the smoothness of function space.
In this case, the convergence rate in~\eqref{excess_risk} becomes quite slow in $n$. 
Nonetheless, high-dimensional data are often observed in real-world applications. 
For instance, in ImageNet challenge, data are RGB images with a resolution of $224 \times 224$, which means $d=3 \times 224 \times 224$. 
Then, the rate in~\eqref{excess_risk} cannot explain the empirical success of deep learning. 
Motivated by this, many researchers have put in considerable effort to avoid the $d$-dependence in the denominator of the rate in~\eqref{excess_risk}.
\\ \\
Several routes exist to achieve the goal: 
One typical approach is to consider the various types of function spaces $\mathcal{G}$ with various smoothness: mixed-Besov space~\citep{suzuki2018adaptivity}, Korobov space~\citep{montanelli2019new}. 
Another alternative is to impose structural assumption on target function $f_{\rho}$.
Such structures include additive ridge functions~\citep{fang2022optimal}, composite functions with hierarchical structures~\citep{schmidt2020nonparametric,han2020depth}, generalized single index model~\citep{bauer2019deep} or multivariate adaptive regression splines (MARS)~\citep{kohler2022estimation}.
Another line of work focuses on the geometric structure of the feature space $\mathcal{X}$. 
These works take the advantage of high-dimensional data having practically low-intrinsic dimensionality~\citep{tenenbaum2000global,roweis2000nonlinear}.
Under this setting,~\cite{nakada2020adaptive} showed deep neural networks can achieve a fast rate over a broad class of measure on $\mathcal{X}$, such as data on highly non-smooth fractal sets. 
This should be contrasted with the fact that linear estimators, which are known to be adaptive to intrinsic dimensions, can achieve fast convergence rates only when the data lie on smooth manifolds.
Another work,~\cite{chen2022nonparametric}, showed the adaptiveness of deep ReLU networks to the data with low-dimensional geometric structures. 
They were interested in estimating target function $f_{\rho}$ in H\"older spaces defined over a low-dimensional manifold $\mathcal{M}$ embedded in $\mathbb{R}^{d}$.
Figure~\ref{Fig0} summarizes the cases on classes $\mathcal{G}$ where neural nets exhibit superior adaptabilities over the classical estimators.
\\ \\
Recently, \cite{suh2022approximation} studied deep ReLU networks estimating H\"older functions on a unit sphere and showed that these networks can avoid the curse of dimensionality as function smoothness increases with the data dimension $d$, i.e., $r=\mathcal{O}(d)$. 
This behavior was not observed in the aforementioned literature, where $\mathcal{X}$ is set as a cube, i.e., $\mathcal{X}:=[0,1]^{d}$.
When $\mathcal{X}$ is a cube, several studies~\citep{lu2021deep,jiao2021deep,shen2019deep} have attempted to track and reduce the $d$-dependence in the constant factor hidden in the Big-O notation in~\eqref{excess_risk}. 
Interested readers can find the detailed comparisons of the results from these works in the Appendix~C in~\cite{suh2022approximation}. 
\\ \\
\noindent \textbf{{Remark 1}}: Note that the aforementioned works are based on the constructions of sparse networks. 
From a technical perspective, the sparsity assumption is natural in the sense of the equation (\ref{Decomposition}).
Nonetheless, as mentioned in~\cite{ghorbani2020discussion}, it is still an open question whether the sparsity (i.e., $\mathcal{N}$) is a sufficient complexity measure of $\mathcal{F}$ for generalizability, as densely connected networks are observed more commonly in practice without the regularized penalties. 
These often lead to overparametrized networks with huge complexity on $\mathcal{F}$, not guaranteeing good generalizability in the sense of (\ref{Decomposition}).
Given this observation, one popular heuristic argument for explaining the good generalizability of dense neural nets is the \textbf{\textit{implicit regularization}} of gradient-based algorithms; that is, the model complexity is not captured by an explicit penalty but by the dynamics of the algorithms implicitly.
For some special cases~\citep{gunasekar2018characterizing, ji2018risk}, it has been shown that the gradient-based methods provably find the solutions of low-complexity in the huge parameter space. (e.g., low-rank matrix in matrix estimation problem.)
A similar phenomenon has been empirically observed in function estimation problems via neural networks \citep{cao2019generalization, hu2020surprising}, sparking further research to be discussed in the next section.
For more in-depth discussions on these issues beyond what is covered in this review, readers should consult Section $3$ of~\cite{he2020recent},  \cite{neyshabur2017implicit}, and references therein.
\\ \\
\noindent \textbf{{Remark 2}}: Reviews on approximation-based statistical guarantees of neural networks for \textit{\textbf{classification problems}} are provided in the Appendix~{\color{blue}B} due to page limit. 

\section{Training dynamics-based statistical guarantees} \label{training_dynamics}
The literature introduced in the Section~\ref{Approx_theory} relied on the assumption that the global minimizer of the empirical risk, $\widehat{f}_{n}$ in~\eqref{Minimizer}, is obtainable. 
However, due to the non-convex nature of the loss function, neural networks estimated using commonly employed gradient-based methods lack guarantees of finding $\widehat{f}_{n}$, which leads to the following natural question: 
\begin{center}
    \textbf{\textit{Does the neural network estimated by gradient-based methods generalize well?}}
\end{center}
The papers to be introduced shortly will try to answer to the above question. 
Due to the complex nature of the problem, most papers to be reviewed have considered the following \textit{shallow} neural network (i.e., networks with one hidden layer) $f_{\mathbf{W}}(\mathbf{x})$ with $M$ number of hidden neurons:
\begin{align} \label{Shallow}
    f_{\mathbf{W}}(\mathbf{x}) = \frac{\alpha}{M}\sum_{r=1}^{M}a_{r}\sigma(w_{r}^{\top}\mathbf{x}).
\end{align}
Notation $\mathbf{x} \in \mathcal{X} \subset \mathbb{R}^{d}$ is an input vector, $\{w_{r}\in\mathbb{R}^{d}\}_{r=1}^{M}$ are the weights in the first hidden layer, and $\{a_{r}\in\mathbb{R}\}_{r=1}^{M}$ are the weights in the output layer.
Let us denote the pair $\mathbf{W}:=\{(a_{r}, w_{r})\}_{r=1}^{M}$.
The network dynamic is scaled with the factor $\frac{\alpha}{M}$.
If the network width (i.e., $M$) is small, the scaling factor has negligible effects on the network dynamics.
But for the wide enough networks (i.e., overparametrized setting), the scaling difference yields completely different behaviors in the dynamics.
Given the $M$ is large enough, we will focus on two specific regimes: 
\begin{enumerate}
    \item[1.] Neural Tangent Kernel regime~\citep{jacot2018neural,du2019gradient} with $\alpha=\sqrt{M}$. (Subsection~\ref{NTK_sec})
    \item[2.] Mean Field regime~\citep{mei2018mean,mei2019mean} with $\alpha=1$. (Subsection~\ref{MF_sec})
\end{enumerate}
Additionally, we will review several works that try to address the drawbacks of the NTK framework, as well as some that provide unifying perspectives on these two regimes (Subsection~\ref{Unifying}). 
\\ \\
We focus on reviewing the papers that work on $\ell_{2}$-loss function :
$\mathcal{L}_{\mathbf{S}} \big(\mathbf{W}\big) = \frac{1}{2}\sum_{i=1}^{n} \big( y_{i}-f_{\mathbf{W}}(\mathbf{x}_{i})\big)^{2}$.
Note that in contrast to~\eqref{Minimizer}, structural requirements such as $\mathcal{F}(L,\mathbf{p},\mathcal{N})$ are removed. 
The model parameter pairs $\mathbf{W}$ are updated through the gradient-based methods. 
Let $\mathbf{W}_{(0)}$ be the initialized weight pairs. 
Then, we have the following gradient descent (GD) update rule with step-size $\eta>0$ and $k \geq 1$:
\begin{align} \label{GD_update}
    \textbf{GD :} \quad \mathbf{W}_{(k)} = \mathbf{W}_{(k-1)} - \eta   \nabla_{\mathbf{W}}\mathcal{L}_{\mathbf{S}} \big(\mathbf{W}\big)\mid_{\mathbf{W}=\mathbf{W}_{(k-1)}}.
\end{align}
Another celebrated gradient-based method is Stochastic Gradient Descent (SGD). 
The algorithm takes a randomly sampled subset (i.e., $\mathcal{B}$) of the data $\mathcal{D}$, computes the gradient with the selected samples, and this significantly reduces the computational burdens in GD. 
Another frequently adopted algorithm in practice is Noisy Gradient Descent (NGD), which adds the centered Gaussian noise to the gradient of loss function in~\eqref{GD_update}. 
It is known that adding noises to gradient helps training~\citep{neelakantan2015adding} and generalization~\citep{smith2020generalization} of neural networks.

\subsection{Neural Tangent Kernel Perspective} \label{NTK_sec}
Over the past few years, Neural Tangent Kernel (NTK)~\citep{arora2019exact,jacot2018neural,lee2018deep,chizat2018global} has been one of the most seminal discoveries in the theory of neural networks.
The underpinning idea of the NTK-type theory comes from the observation that in a {\em wide-enough} neural net, 
model parameters updated by gradient descent (GD) stay close to their initializations during the training, so that 
the dynamics of the networks can be approximated by the first-order Taylor expansion with respect to its parameters at initialization; that is, we denote the output of neural network as $f_{\mathbf{W}_{(k)}}(\mathbf{x})\in\mathbb{R}$ with input $\mathbf{x}\in\mathcal{X}$ and model parameter $\mathbf{W}_{(k)}$ updated at $k$-th iteration of gradient descent, then the dynamics of $f_{\mathbf{W}_{(k)}}(\mathbf{x})$ over $k \geq 1$ can be represented as follows:
\begin{align} \label{NTK_Approx}
    f_{\mathbf{W}_{(k)}}(\mathbf{x}) = 
    f_{\mathbf{W}_{(0)}}(\mathbf{x}) + \langle \nabla f_{\mathbf{W}_{(0)}}(\mathbf{x}), \mathbf{W_{(k)}}-\mathbf{W}_{(0)} \rangle + o(\| \mathbf{W_{(k)}}-\mathbf{W}_{(0)} \|_{\text{F}}^{2}),
\end{align}
where $o(\| \mathbf{W_{(k)}}-\mathbf{W}_{(0)} \|_{\text{F}}^{2})$ is the small random quantity that tends to $0$ as network width gets close to infinity, measuring the distance between updated model parameter and its initialization in Frobenius norm.
Specifically, it can be shown $\| \mathbf{W_{(k)}}-\mathbf{W}_{(0)} \|_{\text{F}}^{2} \leq \mathcal{O}(\frac{1}{\sqrt{M}})$ with sufficiently large enough $M$. 
(For instance, see Remark $3.1$ in \cite{du2018gradient}.)
Under this setting, the right-hand side of~\eqref{NTK_Approx} is linear in the network parameter $\mathbf{W}_{(k)}$.
As a consequence, training on $\ell_{2}$-loss with gradient descent leads to kernel regression solution with respect to the (random) kernel induced by the feature mapping $\phi(\mathbf{x}):=\nabla_{\mathbf{W}_0}f(\mathbf{x})$ for all $\mathbf{x}\in\mathcal{X}$. 
The inner-product of two feature mappings evaluated at two data points $\mathbf{x}_{i}, \mathbf{x}_{j}$ is denoted as $\mathbf{K}^{(M)}(\mathbf{x}_{i},\mathbf{x}_{j}):=\langle \phi(\mathbf{x}_{i}), \phi(\mathbf{x}_{j})\rangle$ for all $1\leq i,j \leq n$.
\\ \\
Note that the $\mathbf{K}^{(M)}(\cdot,\cdot)$ is a random matrix with respect to the initializations $\mathbf{W}_{(0)}$.
It is shown that it converges to its deterministic limit (i.e., $M\rightarrow{\infty}$) in probability pointwisely~\citep{jacot2018neural,arora2019exact,lee2018deep} and uniformly~\citep{lai2023generalization} over $\mathcal{X} \times \mathcal{X}$.
The limit matrix is named NTK denoted as $\{\mathbf{K}^{\infty}(\mathbf{x}_{i},\mathbf{x}_{j})\}_{1\leq i,j\leq n}\in\mathbb{R}^{n \times n}$.
Hereafter, we write the eigen-decomposition of $\mathbf{K}^{\infty}=\sum_{j=1}^{n}\lambda_{j}\mathbf{v}_{j}\mathbf{v}_{j}^{\top}$, where $\lambda_{1}\geq\dots\geq\lambda_{n}\geq 0$ with corresponding eigenvectors $\mathbf{v}_{j}\in \mathbb{R}^{n}$.
\\ \\
\noindent \textit{\textbf{{Optimization of neural nets in NTK regime.} }}
Many papers have come out in the sequel to tackle the optimization properties of neural networks in the NTK regime.
Under the above setting,~\cite{du2018gradient} proves the linear convergence of training loss of shallow ReLU networks. 
Specifically, the authors randomly initialized $a_{r}\sim\text{Unif}\{-1,+1\}$ and $\mathbf{w}_{r}\sim\mathcal{N}(0,\mathcal{I})$, and train the $\mathbf{w}_{r}$ via GD with a constant positive step size $\eta=\mathcal{O}\big(1\big)$.
Here, the linear convergence rate means that the training loss at $k$-th gradient descent decays at a geometric rate with 
respect to the initial training loss, which is explicitly stated in Theorem 4.1. in~\cite{du2018gradient} as:
\begin{align} \label{loss1}
    \|f_{\mathbf{W}_{(k)}}(\mathbf{x}) - \mathbf{y}\|_{2}^{2}
    \leq \bigg( 1-\frac{\eta\lambda_{n}}{2} \bigg)^{k}
    \|f_{\mathbf{W}_{(0)}}(\mathbf{x}) - \mathbf{y}\|_{2}^{2}.
\end{align}
Their result requires the network width $M$ to be in the order of $\Omega\big(\frac{n^{6}}{\lambda_{n}}\big)$, and the decay rate is dependent on the minimum eigenvalue of NTK, $\lambda_{n}$.
Here, for the geometric decay rate, $\lambda_{n}$ needs to be strictly greater than $0$ induced from data non-parallel assumption. (i.e., no two inputs are parallel.) 
\\ \\
Afterwards, there have been several attempts to reduce the overparametrization size.
One work we are aware of is~\cite{song2019quadratic} where they used matrix Chernoff bound to reduce the width size up to $M=\Omega\big(\frac{n^{2}}{\lambda_{n}^{4}}\big)$ with a slightly stronger assumption than the data non-parallel assumption. 
Several subsequent works~\cite{allen2018convergence,du2019gradient,zou2018stochastic,wu2019global,oymak2020toward,suh2021non} extended the results showing the linear convergence of training loss of \textit{deep} ReLU networks with $L$-hidden layers.
For a succinct comparison of the overparametrized conditions on $M$ in aforementioned papers, we direct readers to Table 1 in~\cite{zou2019improved}.
\\ \\
\noindent \textit{\textbf{{Spectral bias of shallow ReLU networks.} }}
Motivated by the result~\eqref{loss1}, researchers further studied the {\textit{spectral bias}} of deep neural networks, investigating why the neural dynamics learn the lower frequency components of the functions faster than they learn the higher frequency counterparts.
The specific results are stated in terms of eigenvalues $\mu_{1}\geq\mu_{2}\geq\dots$ and corresponding orthonormal eigenfunctions $\phi_{1}(\cdot), \phi_{2}(\cdot), \dots$ of integral operator $\mathcal{L}_{\mathbf{K}^{\infty}}$ induced by $\mathbf{K}^{\infty}$:
\begin{align*}
    \mathcal{L}_{\mathbf{K}^{\infty}}(f)(\mathbf{x}):=\int_{\mathcal{X}}\mathbf{K}^{\infty}(\mathbf{x},\mathbf{y})f(\mathbf{y})\rho(\mathbf{dy}),
    \quad \forall f\in\mathcal{L}^{2}(\mathcal{X}),
\end{align*}
where $\mathcal{L}^{2}(\mathcal{X})$ is a $L_{2}$-space on $\mathcal{X}$.
Specifically,~\cite{cao2019towards,bietti2019inductive} provided the spectral decay rates of $(\mu_{k})_{k}$ for shallow ReLU networks when $\mathbf{x}$ is from a unit-sphere equipped with uniform measure as follows:~\footnote{Note that eigenvalues of $\mathbf{K}^{\infty}$ (i.e.,$\{\lambda_{i}\}_{i=1}^{n}$) and eigenvalues of $\mathcal{L}_{k}$ (i.e.,$(\mu_{k})_{k}$) are different.}

\begin{proposition}
    (Theorem $4.3$ in~\cite{cao2019towards}, Proposition 5 in~\cite{bietti2019inductive}) For the neural tangent kernel corresponding to a two-layer feed-forward ReLU network, the eigenvalues $(\mu_{k})_{k}$ satisfy the following:
    \begin{align*}
        \begin{cases}
            \mu_{k} = \Omega(1),  \quad &\text{when $k=0, 1$},  \\
            \mu_{k} = 0,  \quad &\text{when $k (\geq 3)$ is odd},  \\
            \mu_{k} = \Omega(\text{max}(k^{-d-1},d^{-k-1})),  \quad &\text{when $k (\geq 2)$ is even}. \\
        \end{cases}
    \end{align*}
\end{proposition}
\noindent
The decay rate is exponentially fast in input dimension $d$ for $k \gg d$.
An interesting benefit of having a specific decay rate is that we can measure the size of reproducing kernel Hilbert spaces (RKHS) induced from the kernel $\mathbf{K}^{\infty}$. (We denote this RKHS as $\mathcal{H}^{\infty}$ for the later use.)
The slower the decay rate is, the larger the RKHS becomes, allowing higher-frequency information of function to be included. 
\\ \\
With the specified eigendecay rates on $\mu_{k}$, Theorem $4.2$ in~\cite{cao2019towards} proved the spectral bias of neural network training in the NTK regime.
Specifically, as long as the network is wide enough and the sample size is large enough, gradient
descent first learns the target function along the eigen directions of NTK with larger
eigenvalues, and learns the rest components corresponding to smaller eigenvalues later. 
Similary,~\cite{hu2019simple} showed that gradient descent learns the linear component of target functions in the early training stage.
But crucially they do not require the network to have a disproportionately large width, and the network is allowed to escape the kernel regime later in training. 
\\ \\
\noindent \textit{\textbf{{Generalization of neural nets in NTK regime.} }}
Here, we review some important works that study the generalizabilities of~\eqref{Shallow}.
To the best of our knowledge,~\cite{arora2019fine} provided the first step in understanding the role of NTK in the generalizability of neural nets.
Specifically, they showed that for $M=\Omega(\frac{n^{2}\log(n)}{\lambda_{n}})$ and $k\geq\Tilde{\Omega}\big(\frac{1}{\eta \lambda_{n}}\big)$, the $\ell_{2}$-population loss of $f_{\mathbf{W}_{(k)}}(\mathbf{x})$ is bounded by:
\begin{align} \label{NTK_GEN}
    \mathbb{E} \big[ (f_{\mathbf{W}_{(k)}}(\mathbf{x})-\mathbf{y})^{2}\big] \leq 
    \mathcal{O}\bigg(\sqrt{\frac{\mathbf{y}^{\top}(\mathbf{K}^{\infty})^{-1}\mathbf{y}}{n}}\bigg).
\end{align}
Observe the nominator in the bound can be written as $\mathbf{y}^{\top}(\mathbf{K}^{\infty})^{-1}\mathbf{y}:=\sum_{i=1}^{n}\frac{1}{\lambda_{i}}(\mathbf{v}_{i}^{\top}\mathbf{y})^{2}$.
This implies the projections $\mathbf{v}_{i}^{\top}\mathbf{y}$ that correspond to small eigenvalues $\lambda_{i}$ should be small for good generalizations on unseen data.
This theoretical result is consistent with the empirical finding in~\cite{zhang2017discrimination}. 
Indeed, they performed their own empirical experiments on MNIST and CIFAR-10 datasets, showing the projections $\{(\mathbf{v}_{i}^{\top}\mathbf{y})\}_{i=1}^{n}$ sharply drops for true labels $\mathbf{y}$, leading to the fast convergence rate.  In contrast, when the projections are close to being uniform for random labels $\mathbf{y}$, leading to slow convergence.
(See Figure $1$ in~\cite{zhang2017discrimination})
\\ \\
However, the bound~\eqref{NTK_GEN} is obtained in the noiseless setting and becomes vacuous under the presence of noise.
In this regard, under the noisy setting~\eqref{non_para}, \cite{nitanda2020optimal} showed that 
\begin{align}\label{SUZUKI}
    \mathbb{E}\big[\|f_{W_{(k)}}(\mathbf{x})-f_{\rho}(\mathbf{x})\|_{L_{2}}^{2}\big] \leq \mathcal{O}\big(k^{-\frac{2r\beta}{2r\beta+1}}\big),
\end{align}
where the target function $f_{\rho}$ belongs to the subset of $\mathcal{H}^{\infty}$, and $f_{W_{(k)}}$ is a shallow neural network with a smooth activation function that approximates ReLU. 
Here, the network is estimated via one-pass SGD (take one sample for gradient update and the samples are visited only once during training), minimizing $\ell_{2}$-regularized expected loss.
This setting leads to $k=n$.
The rate~\eqref{SUZUKI} is minimax optimal, which is faster than $\mathcal{O}(\frac{1}{\sqrt{n}})$ in~\cite{arora2019fine}.
It is characterized by two control parameters, $\beta$ and $r$, where $\beta>1$ controls the size of $\mathcal{H}^{\infty}$ and $r\in[1/2,1]$ controls the size of subset of $\mathcal{H}^{\infty}$ where $f_{\rho}$ belongs. 
The bound~\eqref{SUZUKI} has an interesting {bias-variance} trade-off between these two quantities $\beta$ and $r$.
For large $\beta$, the whole space $\mathcal{H}^{\infty}$ becomes small, and the subspace of $\mathcal{H}^{\infty}$ needs to be as large as possible for the faster convergence rate, and vice versa.
\\ \\
However, as noted by the following work~\citep{hu2021regularization}, the rate in~\eqref{SUZUKI} requires the network width $M$ to be exponential in $n$.
The work reduced the size of overparametrization to $\Tilde{\Omega}(n^6)$ when the network parameters are estimated by gradient descent. 
The paper proved that the overparametrized shallow ReLU networks require $\ell_{2}$-regularization for GD to achieve minimax convergence rate $\mathcal{O}(n^{-\frac{d}{2d-1}})$.
Afterward,~\cite{suh2021non} extended the result to \textit{deep} ReLU networks in the NTK regime, showing that $\ell_{2}$-regularization is also required for achieving minimax rate for deep networks.

\subsection{Mean-Field Perspective} \label{MF_sec}
A \textit{``mean-field''} viewpoint is another interesting paradigm to help us understand the optimization landscape of neural network models.
Recall that neural network dynamics in the Mean-Field (MF) regime corresponds to $\alpha=1$ in~\eqref{Shallow}. 
\\ \\
The term \textit{mean-field} comes from an analogy with mean field models in mathematical physics, which analyze the stochastic behavior of many identical particles~\citep{Lenya2023note}.  
Let us denote $\theta_{r}:=(a_{r},w_{r})\in\mathbb{R}^{d+1}$,
$\sigma_{\star}(\mathbf{x},\theta_{r}):=a_{r}\sigma(w_{r}^{\top}\mathbf{x})$ in~\eqref{Shallow}.
Weight pairs $\{\theta_{r}\}_{r=1}^{M}$ are considered as a collection of gas particles in $\mathbf{D}$-dimensional spaces with $\mathbf{D}:=d+1$.
We consider there are infinitely many gas particles allowing $M\rightarrow{\infty}$ which yields the following integral representation of neural dynamics:
\begin{align} \label{Shallow-mean-field}
    \frac{1}{M}\sum_{r=1}^{M}\sigma_{\star}(\mathbf{x};\mathbf{\theta}_{r})
    \xrightarrow{M\rightarrow\infty}
    f(\mathbf{x};\rho) := \int \sigma_{\star}(\mathbf{x};\mathbf{\theta})\rho(d\mathbf{\theta}),
\end{align}
where $\theta_{r}\sim\rho$ for $r=1,\dots,M$.
The integral representation~\eqref{Shallow-mean-field} is convenient for the mathematical analysis as it is linear with respect to the measure $\rho$.
For instance, see~\cite{bengio2005convex}.
\\ \\
Under this setting, the seminal work~\citep{mei2018mean} studied the evolution of particles $\theta^{(k)}\in\mathbb{R}^{\mathcal{D}}$ updated by $k$-steps of one-pass SGD (take one sample for gradient update and the samples are visited only once during training) under $\ell_{2}$-loss.
Interestingly, they proved the trajectories of empirical distribution of $\theta^{(k)}$ denoted as $\widehat{\rho}_{k}:=\frac{1}{M}\sum_{r=1}^{M}\delta_{\theta^{(k)}_{r}}$ weakly converges to the deterministic limit $\rho_{t}\in\mathcal{P}(\mathbb{R}^{\mathcal{D}})$ as $k\rightarrow{\infty}$ and $M\rightarrow{\infty}$.
The measure $\rho_{t}$ is the solution of the following nonlinear partial differential equation (PDE):
\begin{align} 
   &\partial_{t}\rho_{t}= \nabla_{\mathbf{\theta}}\cdot\big(\rho_{t} \nabla_{\mathbf{\theta}} \mathbf{\Psi}(\mathbf{\theta};\rho_{t})\big),
    \qquad \mathbf{\Psi}(\mathbf{\theta};\rho_{t}):=\mathcal{V}(\mathbf{\theta}) + \int\mathcal{U}(\mathbf{\theta},\Bar{\mathbf{\theta}})\rho_{t}(d\Bar{\mathbf{\theta}}), \label{PDE_main} \\
    &\mathcal{V}(\mathbf{\theta}):=-\mathbb{E}\{\mathbf{y}\sigma_{\star}(\mathbf{x};\mathbf{\theta})\},
    \qquad \qquad \mathcal{U}(\mathbf{\theta_{1}},\mathbf{\theta_{2}}):=\mathbb{E}\{\sigma_{\star}(\mathbf{x};\mathbf{\theta}_{1})\sigma_{\star}(\mathbf{x};\mathbf{\theta}_{2})\}. \nonumber
\end{align}

The above PDE describes the evolution of each particle $(\theta_{r})$ in the force field created by the densities of all the other particles. (We provide more descriptions on the above PDE in the Appendix~{\color{blue}C}.) 
Denote $\mathcal{R}(\rho_{t}):=\mathbb{E}[(y-f(\mathbf{x};\rho_{t}))^{2}]$ and let $\mathcal{R}_{M}$ be the empirical version of $\mathcal{R}$, then under some regularity assumptions on the network, we have: 
\begin{align}\label{MF_bound}
    \sup_{0\leq t\leq T}\big| \mathcal{R}(\rho_{t})-\mathcal{R}_{M}(\widehat{\rho}_{\floor{t/2\eta}})\big|
    \leq e^{C(T+1)}\cdot \sqrt{\frac{1}{M} \vee 2\eta } \cdot \sqrt{D+\log\frac{M}{2\eta}}.
\end{align}
\noindent
The condition for the bound to vanish to $0$ is 
(1) $M \gg D$, (2) $\eta \ll \frac{1}{D}$, and (3) PDE converges in $T=\mathcal{O}(1)$ iterations. 
It is interesting to note that the generic ODE approximation requires the step-size $\eta$ to be less than the order of the total number of parameters in the model (i.e., $\eta \ll \frac{1}{MD}$), whereas in this setting the step-size $\eta\ll\frac{1}{D}$ should be enough.
Also, recall that the number of sample size $n$ is equivalent to the iteration steps $k:=\floor{\frac{T}{\varepsilon}}$ of one-pass SGD with $T=\mathcal{O}(1)$. 
Then, this means $n=\mathcal{O}(D) \ll \mathcal{O}(MD)$ should be enough for a good approximation. 
Another notable fact is; in contrast to the NTK regime, the evolution of weights $\theta_{r}$ are non-linear, and in particular, the weights move away from their initialization during training. 
Indeed under mild assumption, we can show that for small enough step size $\eta$, $\lim_{M\rightarrow\infty}\|\theta^{(k)}-\theta^{(0)} \|_{2}^{2}/M=\Omega(\eta^{2})$ in the mean-field regime, while $\sup_{t\geq 0}\|\theta^{(t)}-\theta^{(0)}\|_{2}^{2}/M=\mathcal{O}(n/(Md))$ in the NTK regime. See~\cite{bartlett_montanari_rakhlin_2021}.
\\ \\
\begin{table}[!t]
\centering
\begin{tabular}{|c|l|l|}
\hline
\multicolumn{1}{|l|}{} & \multicolumn{1}{c|}{\textbf{NTK Regime}}                                                                                                  & \multicolumn{1}{c|}{\textbf{MF Regime}}                                                                                                                                    \\ \hline
\textbf{Pros}                  & \begin{tabular}[c]{@{}l@{}}1. Same scaling as in practice\\ 2. Finite time convergence rate\\ 3. Generalization bounds\end{tabular} & \begin{tabular}[c]{@{}l@{}}1. Does not require $\theta^{(k)}$ to be close to $\theta^{(0)} $\\ 2. Potentially learn a larger class of functions\end{tabular} \\ \hline
\textbf{Cons}                  & Require $\theta^{(k)}$ to be close to $\theta^{(0)}$                                                                                & \begin{tabular}[c]{@{}l@{}}1. Not the same scaling as in practice\\ 2. No finite-time convergence rate\\ 3. No generalization bounds\end{tabular} \\
\hline
\end{tabular}
\caption{NTK v.s. Mean-Field regimes}
\label{Table1}
\end{table}
Despite the nice characterizations of SGD dynamics, still, the bound in~\eqref{MF_bound} has room for the improvement; the number of neurons $M$ is dependent on the ambient data dimension $d$ and the bound is only applicable to the SGD with short convergence iterations $T=\mathcal{O}(1)$.
A follow-up work~\citep{mei2019mean} has attempted to tackle these challenges.
Particularly, they proved that there exists a constant $K$ that only depends on intrinsic features of the activation and data distribution, such that with high probability, the following holds: 
\begin{align}
    \sup_{0\leq t\leq T}\big| \mathcal{R}(\rho_{t})-\mathcal{R}_{M}(\widehat{\rho}_{\floor{t/\varepsilon}})\big|
    \leq
    K e^{K (T\eta)^{3}}\bigg\{ \sqrt{\frac{\log(M)}{M}}+\sqrt{d+\log(M)}\sqrt{\eta}\bigg\}.
\end{align}
A remarkable feature of this bound is that as long as $T\eta=\mathcal{O}(1)$ and $K=\mathcal{O}(1)$, the number of neurons only needs to be chosen $M\gg 1$ for the mean-field approximation to be accurate.
The condition $T\eta=\mathcal{O}(1)$ mitigates the exponential dependence on $T$ and the bound does not need to scale with the ambient dimension $d$.
Later, researchers from the same group generalize the result into multi-layer settings~\citep{nguyen2020rigorous}. 

\subsection{Beyond the NTK and Mean-Field regimes} \label{Unifying}
Despite nice theoretical descriptions on training dynamics of gradient descent in loss functions,~\cite{arora2019exact, lee2018deep, chizat2018global} empirically found significant performance gaps between NTK and actual training in many downstream tasks. 
For instance,~\cite{arora2019exact} derived the convolution neural network (CNN)-based CNTK, and empirically found the $5\% \sim 6\%$ performance gaps between CNN and CNTK-based kernel regressor in image classification tasks. (i.e., CNN performs better.)
This indicates the potential benefits of the finite width in neural networks. 
\\ \\
\noindent \textit{\textbf{{Beyond the NTK regime.} }}
These gaps have been theoretically studied in several papers, including \cite{wei2019regularization, allen2019can, ghorbani2020neural, yehudai2019power, chizat2018global}. 
They showed that NTK has worse generalization guarantees than finite-width neural networks in some settings. For example, \cite{mei2018landscape} demonstrated that training a neural network with one hidden neuron can efficiently learn a single neuron target function with $\mathcal{O}(d \log d)$ samples, whereas the corresponding RKHS has a test error that is bounded away from zero for any sample size polynomial in $d$ \citep{yehudai2019power, ghorbani2021linearized}. 
However, kernel methods often perform comparably to neural networks in some image classification tasks \citep{li2019enhanced, novak2018bayesian}. 
\cite{ghorbani2020neural} provided a unified framework under spiked covariate models to explain this divergence, showing that while RKHS and neural networks perform similarly in certain stylized tasks, RKHS performance deteriorates under isotropic covariate distributions (e.g., noisy high-frequency image components), whereas neural networks are less affected by such noise.
Another work,~\cite{wei2019regularization} gives an interesting example of where NTK or any kernel methods are statistically limited, whereas regularized neural networks have better sample complexity. 
They proved that there is a $\Omega(d)$ sample-complexity gap between the regularized neural net and kernel prediction function for estimating $f_{\rho}(\mathbf{x})=\mathbf{x}_{1}\mathbf{x}_{2}$ with $\mathbf{x}_{i}\sim\{\pm 1\}$ for $\mathbf{x}\in\mathbb{R}^{d}$.
\\ \\
The aforementioned works explained the superiority of neural networks over the networks in the NTK regime under some \textit{highly stylized} settings.
There also has been another line of works~\citep{li2018learning,allen2019learning,bai2019beyond} to explain how the networks estimated through gradient-based methods generalize well but critically do not rely on the linearization of network dynamics.
Under the distribution-free setting (i.e., no distributional assumptions on covariates), \cite{allen2019learning} provided optimization and generalization guarantees for three-layer ReLU networks learning the function classes of three-layer networks with smooth activation functions. They showed that three-layer ReLU networks can learn a larger function class than two-layer ReLU networks do. Unlike NTK techniques, their approach allows non-convex interactions across hidden layers, enabling parameters trained by SGD to move far from their initializations.
Motivated from~\cite{allen2019learning}, the paper~\citep{bai2019beyond} studied the optimization and generalization of shallow networks with smooth activation function $\sigma(\cdot)$ via relating the network dynamics $f_{\mathbf{W}}(\mathbf{x})$ in~\eqref{Shallow} with second-order (or quadratic) approximations. 
They explicitly showed the sample complexity of the quadratic model is smaller than that of the linear NTK model in learning some stylized target functions by the factor of $\mathcal{O}(d)$.
Similarly, relying on the tensor decomposition techniques instead of quadratic approximation, \cite{li2018learning} showed the separations of shallow ReLU networks and NTK regressors. 
\begin{figure}[!t]
  \centering
  \includegraphics[width=1\textwidth]{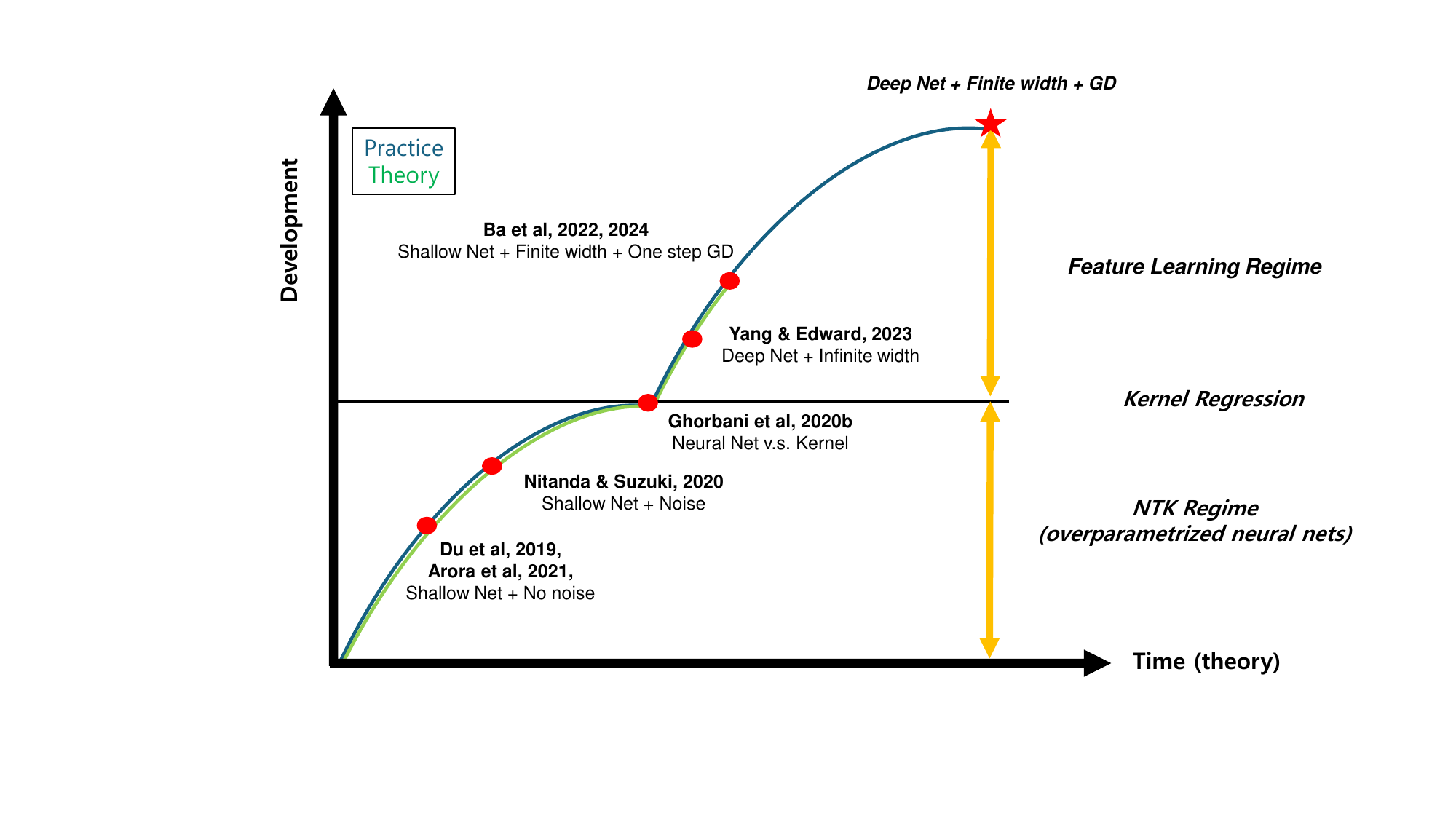}
  \caption{Development of the literature (y-axis) on algorithm-based neural network analysis over time (x-axis). We view the ultimate goal (represented as star) of this line of research is to theoretically demystify feature learning of neural nets with deep layers and finite width, closing the gap with the practice. Note that kernel regressor in NTK regime does not exhibit feature learning functionality.}
  \label{Fig1}
\end{figure}
\\ \\
\noindent\textit{\textbf{{Unifying views of NTK and MF regimes.}}}
 There have been several attempts to give a unifying view of the two regimes; NTK and Mean-Field. 
An attempt we are aware of is~\cite{chen2020generalized}.
The motivation of this paper is summarized in Table~\ref{Table1}, along with the pros and cons of NTK and Mean-Field regimes. 
They showed the two-layer neural networks learned through noisy-gradient descent (NGD) in MF scaling can potentially learn a larger class of functions than networks in NTK scaling can do.
One seminal work, \cite{yang2020feature}, identifies a set of scales for initialized weights and SGD step sizes where \textbf{feature learning} occurs in \textbf{deep} neural networks in the infinite width limit. 
They offer a unified framework that encompasses parameterizations in both the NTK and MF regimes.
Feature learning is the core property of neural networks: \textit{the ability to learn useful features} out of raw data~\citep{girshick2014rich,devlin2018bert} that adapt to the learning problem.
For instance, BERT~\citep{devlin2018bert} leveraged this property of neural networks for sentence sentiment analysis.
Specifically, in the regime where the network width and data size are comparable,~\cite{ba2022high} showed the non-trivial feature learning occurs at the early phase (one gradient step in GD) of shallow neural network training with a large enough step-size. 
Similarly to \cite{ghorbani2020neural}, \cite{ba2024learning} examined the advantages of shallow neural networks with finite width over kernel methods under the spiked covariance model. Both studies demonstrated that neural networks and kernel methods benefit from stronger low-dimensional structures (i.e., larger spikes). However, \cite{ba2024learning} focused on gradient-based optimization guarantees, while \cite{ghorbani2020neural} provided only approximation-based analysis.
\\ \\
Interested readers can find more detailed descriptions on the works~\cite{wei2019regularization, ghorbani2020neural, bai2019beyond, chen2020generalized, yang2020feature} in the  Appendix~{\color{blue}D} and~{\color{blue}E}. 
In Figure~\ref{Fig1}, we summarizes our own views on the developments of literatures along this line of research. 

\section{Statistical Guarantees of Generative Models} \label{Generative_Models}
In this section, we sequentially explore the statistical literature on three topics: Generative Adversarial Networks (GANs), diffusion models, and in-context learning in large language models (LLMs).

\subsection{Generative Adversarial Networks (GAN)}
Over the past decade, Generative Adversarial Networks (GANs)~\citep{goodfellow2014generative} have stood out as a significant unsupervised learning approach, known for their ability to learn 
the data distribution and efficiently sample the data from it.
The main goal of GAN is to learn the target distribution $\mathbf{X} \sim \nu$ through the adversarial training between a \textit{Discriminator} and a \textit{Generator}.
Here, the generator takes the input $\mathbf{\mathcal{Z}}$ from prior distributions such as Gaussian or Uniform (i.e., $\mathcal{Z}\sim \pi$) and the input is push-forwarded by transformation map $g:\mathbf{\mathcal{Z}} \rightarrow g(\mathbf{\mathcal{Z}})$.
In this seminal paper, the distribution of random variable $g(\mathcal{Z})$ is referred to as an {implicit} distribution $\mu$.
The primary objective of the generator is to produce synthetic data from $\mu$ that closely resembles samples from the target distribution.
\\ \\
The adversarial training between the discriminator and generator is enabled through optimizing the following minimax problem w.r.t. functions in the generator class $\mathcal{G}$ and discriminator class $\mathcal{F}$:
\begin{align} \label{GAN}
    (g^{\star}, f^{\star}) \in \argmin_{g\in\mathcal{G}}
    \argmax_{f\in\mathcal{F}}\bigg\{ \mathbb{E}_{\mathcal{Z}\sim\pi} f(g(\mathcal{Z})) - \mathbb{E}_{X\sim\nu} f(X) \bigg\}.
\end{align}
In practice, the above expectations can be approximated with $m$ training data from $\nu$ and $n$ drawn samples from $\mu$.
The inner maximization problem in~\eqref{GAN} is an Integral Probability Metric (IPM, ~\cite{muller1997integral}), which quantifies the discrepancy between two distributions $\mu$ and $\nu$ w.r.t. a symmetric function class, i.e., $f\in\mathcal{F}$, then $-f\in\mathcal{F}$;
\begin{align} \label{IPM}
    d_{\mathcal{F}}(\mu, \nu) = \sup_{f\in\mathcal{F}} \bigg\{ \mathbb{E}_{x\sim \mu}f(x) - \mathbb{E}_{y \sim \nu}f(y) \bigg\}.
\end{align}
When $\mathcal{F}$ is taken to be all $1$-Lipschitz functions, $d_{W_{1}}(\cdot,\cdot)$ is the Wasserstein-$1$ distance; when $\mathcal{F}$ is the class of all indicator functions, $d_{\text{TV}}(\cdot, \cdot)$ is the total variation distance; 
when $\mathcal{F}$ is taken as a class of neural networks, $d_{\text{NN}}(\cdot,\cdot)$ is the ``neural net distance'' (see~\cite{arora2017generalization}).
Under this setting, a generator from $\mathcal{G}$ attempts to minimize the IPM between $\mu$ and $\nu$.
\\ \\
\noindent \textit{\textbf{{Generalization of GANs.}}}
A question that naturally arises is: ``What does it mean for GANs to generalize effectively?''
The paper~\cite{arora2017generalization} provided a mathematical definition for generalization in GANs in terms of IPM.
\begin{definition}
    Let $\widehat{\mu}_{n}$ and $\widehat{\nu}_{m}$ be the empirical distributions of $\mu$ and $\nu$.
    For some generalization gap $\varepsilon > 0$, if the following holds with high-probability;
    \begin{align}
        \left| d_{\mathcal{F}}(\mu,\nu)-d_{\mathcal{F}}(\widehat{\mu}_{n},\widehat{\nu}_{m}) \right|\leq \varepsilon,
    \end{align}
    with $n$ being polynomially dependent in $\varepsilon$, the divergence $d_{\mathcal{F}}(\cdot,\cdot)$ between distributions generalizes.
\end{definition}

This means that if the absolute discrepancy between population divergence and empirical divergence of $\mu$ and $\nu$ can be arbitrarily controlled with $n$ polynomially generated samples, GAN generalizes well.
The same paper proved under this definition, that GAN cannot generalize w.r.t. Wasserstein-1 distance and Jensen-Shannon divergence as $n=\Tilde{\mathcal{O}}(\varepsilon^{-\text{poly(d)}})$ is required.
But it generalizes well w.r.t. neural net distance with $n=\Tilde{\mathcal{O}}(p\log(L)\cdot\varepsilon^{-2})$ where $p$ is the total number of parameters in discriminator neural network and $L$ is a Lipschitz constant of discriminators w.r.t. parameters.
\\ \\
Nonetheless, as noted by~\cite{chen2020distribution,zhang2017discrimination,arora2017generalization}, this result has some limitations:
$(1)$ the sample complexity is involved with unknown Lipschitz constants $L$ of discriminator.
$(2)$ A small neural net distance does not necessarily mean that two distributions are close. (Section $3.4$ in~\cite{arora2017generalization})
$(3)$ Sample complexity is not involved with the complexity of generator class $\mathcal{G}$ under the assumption that the generator can approximate well enough the target data distribution.  
$(4)$ No concrete architecture of discriminator networks is given.
Some papers~\cite{zhang2017discrimination,jiang2018computation,bai2018approximability} attempted to address the first two limitations, and their attempts are nicely summarized in~\cite{chen2020distribution}.
In this review, we introduce works~\cite{chen2020distribution, liang2021well} that tackle the issues raised in $(3)$ and $(4)$ concretely through the tools from the approximation theory of neural networks.
\\ \\
\noindent \textit{\textbf{{Statistical guarantees of GANs.}}}
Note that the functions in discriminator and generator classes can be either classical non-parametric regressors such as random forests, local polynomial, etc., or can be both parametrized by neural networks.
Here, we focus on the latter case which is more commonly used in practice. 
Specifically, let us denote $\mathcal{F}:=\{f_{\omega}(\cdot):\mathbb{R}^{d}\rightarrow\mathbb{R}\}$ as the {Discriminator} class and $\mathcal{G}:=\{g_{\theta}(z):\mathbb{R}^{m}\rightarrow\mathbb{R}^{d}\}$ (with $m\leq d$) as the {Generator} class with $\omega$ and $\theta$ being denoted as network parameters of respective classes, and we are interested in estimating the parameters by minimizing following optimization problem: 
\begin{align} \label{GAN_2}
    (\widehat{\theta}_{m,n}, \widehat{\omega}_{m,n}) \in \argmin_{\theta:g_{\theta}\in\mathcal{G}}
    \max_{\omega:f_{\omega}\in\mathcal{F}}\bigg\{ \widehat{\mathbb{E}}_{m} f_{\omega}(g_{\theta}(\mathcal{Z})) - \widehat{\mathbb{E}}_{n} f_{\omega}(X) \bigg\},
\end{align}
where $\widehat{\mathbb{E}}_{m}(\cdot)$ (resp. $\widehat{\mathbb{E}}_{n}(\cdot)$) denotes the empirical expectation with $m$ generator samples. (resp. $n$ traning samples.) 
\\ \\
\textit{\textbf{{Summary of~\cite{liang2021well}}}}
Given that the optimal parameters of generator $\widehat{\theta}_{m,n}$ in~\eqref{GAN_2} can be obtained, \cite{liang2021well} studied how well the \textit{implicit} distribution estimator $\mu_{\widehat{\theta}_{m,n}}$ (i.e., distribution of random variable $g_{\widehat{\theta}_{m,n}}(Z)$) is close to the target distribution $\nu$ in a Total Variation distance.
Under some regularity assumptions on architectures of $g_{\theta}$ and $f_{w}$, Theorem $19$ in~\cite{liang2021well} proved the existence of $(g_{\theta}(z), f_{\omega})$ pairs satisfying the bound:
\begin{align} \label{conv_GAN}
    \mathbb{E} d_{\text{TV}}^{2}(\nu, \mu_{\widehat{\theta}_{m,n}})
    &\leq 
    \sqrt{d^{2}L \log(dL)
    \bigg(\frac{\log m}{m} \vee \frac{\log n}{n}\bigg)}.
\end{align}
In the rate~\eqref{conv_GAN}, $L$ and $d$ are the depth and width of the generator networks, respectively.
This result allows the very deep network as $L\leq \sqrt{(n\wedge m)/\log(n\vee m)}$. 
It is worth noting that the generator requires the width of the network to be the same as the input dimension $d$ so that the invertibility condition on the generator is satisfied.
As for the discriminator, it can be constructed by concatenating two networks that have the same architecture as the one from the generator, and with the additional two layers, i.e., network $f_{\omega}$ has $L+2$ layers.
The $g_{\theta}$ and $f_{\omega}$ used leaky ReLU and dual leaky ReLU as activation functions respectively for their invertibility.
However, this invertibility condition is often violated in practical uses of GANs.
\\ \\
\noindent\textit{\textbf{{Summary of~\cite{chen2020distribution}}}}
A subsequent work~\cite{chen2020distribution} provides more flexible network architectures for $g_{\theta}$ and $f_{\omega}$ without requiring the invertibility condition on generator and activation function, i.e., the authors consider ReLU activation function.
The paper mainly focuses on three interesting scenarios where they impose structural assumptions on target distribution $\nu$:
\begin{enumerate}
    \item The target distribution $\nu$ is assumed to have a $\alpha(>0)$-H\"older density $p_{\nu}$ with respect to Lebesgue measure in $\mathbb{R}^{d}$, and the density is lower-bounded away from $0$ on a compact convex subset $\mathcal{X}\subset\mathbb{R}^{d}$.
    
    \item The target distribution $\nu$ is supported on the $q$-dimensional (i.e., $q\ll d$) linear subspace of the data domain $\mathcal{X}\subset\mathbb{R}^{d}$, where the density function is assumed to be in $\alpha$-H\"older class.
    
    \item The target distribution $\nu$ is supported on $\mathcal{X}\subset [0,1]^{d}$ being the $q$-dimensional (i.e., $q\ll d$) nonlinear $K$-mixture components, where each component's density function is in $\alpha$-H\"older class.
\end{enumerate}
In scenario 1, the discriminator class $\mathcal{F}$ is assumed to be $\beta$-H\" older class for $\beta>1$.
In scenarios 2 and 3, $\mathcal{F}$ is considered to be a collection of $1$-Lipschitz functions.
The convergence rate of IPM, the depth $L$ and max-widths $\mathbf{p}_{\text{max}}$ of the generator and discriminator in each scenario are summarized in Table~\ref{Table2}.

\begin{table}[!t]
\centering
\begin{adjustbox}{width=0.85\textwidth}
\begin{tabular}{|cc|cc|c|}
\hline
                                       &  & $L$ & $P_{\text{max}}$ & Conv. Rate              \\ \hline
\multicolumn{1}{|l|}{\multirow{2}{*}{\textbf{Scenario 1.}}} & $g_{\theta}$ & $\mathcal{O}\big( \frac{\beta}{2\beta + d}\log n \big)$  & $\mathcal{O}\big( d n^{\frac{\beta d}{(\alpha+1)(2\beta+d)}} \big)$ & \multirow{2}{*}{$\Tilde{\mathcal{O}}\big( n^{-\frac{\beta}{2\beta+d}} \log^{2}n \big)$ } \\ \cline{2-2}
\multicolumn{1}{|l|}{}                  & $f_{\omega}$ & $\mathcal{O}\big( \frac{\beta}{2\beta + d}\log n \big)$  & $\mathcal{O} \big( n^{\frac{2d}{2\beta + d}} \big)$  &                   \\ \hline
\multicolumn{1}{|l|}{\multirow{2}{*}{\textbf{Scenario 2.}}} & $g_{\theta}$ & $\mathcal{O}\big( \frac{\alpha}{2\alpha+q}\log n \big)$  & $\mathcal{O}\big( q n^{\frac{q\alpha}{(\alpha+1)(2\alpha+d)}} \vee d \big)$  & \multirow{2}{*}{$\Tilde{\mathcal{O}}\big( n^{-\frac{1}{2+q}} \log^{2}n \big)$ } \\ \cline{2-2}
\multicolumn{1}{|l|}{}                  & $f_{\omega}$ & $\mathcal{O}\big( \frac{1}{2+q}\log n \big)$  & $\mathcal{O}\big( n^{\frac{q}{2+q}}\vee d \big)$  &                   \\ \hline
\multicolumn{1}{|l|}{\multirow{2}{*}{\textbf{Scenario 3.}}} & $g_{\theta}$ &  $\mathcal{O}\big( \frac{1}{q} \log n \big)$ & $\mathcal{O}\big( Kd n^{\frac{1}{\alpha}} \big)$  & \multirow{2}{*}
{$\mathcal{O}\big( d n^{-\frac{1}{q}} \big)$ } \\ \cline{2-2}
\multicolumn{1}{|l|}{}         & $f_{\omega}$ & $\mathcal{O}\big( \log n + d \big)$  &  $\mathcal{O}\big( n^{\frac{d}{q}} \big)$ &     \\        \hline
\end{tabular}
\end{adjustbox}
\caption{Summary of depth (i.e., $L$) and width (i.e., $\mathbf{p}_{\text{max}}$) of generators (i.e., $g_{\theta}$) and discriminators (i.e., $f_{w}$), and the convergences of IPM under three specially-designed scenarios in GAN framework.} 
\label{Table2}
\end{table}

\noindent
In the first scenario, the convergence rate cannot avoid the exponential dependence on $d$, aligning with the known minimax lower bound~\citep{tang2022minimax}.
The result in the second scenario indicates GANs can avoid the curse of dimensionality being adaptive to the unknown low-dimensional linear structures, achieving faster rates than the parametric rate $n^{-\frac{1}{2}}$.
However, rather than the real-world data being centered in the low-dimensional linear subspace, mixture data are more commonly observed in practice, i.e., MNIST data or images in CIFAR-10. 
In the third scenario, the rate depends linearly on $d$ and exponentially on $q$, showing GANs can capture nonlinear data structures. 
The depth $L$ of networks grows logarithmically with sample size $n$. 
In contrast, to~\cite{liang2021well}, the widths of networks are not the same with the input dimension $d$.

\subsection{Score-based diffusion models}
Score-based diffusion model consists of two processes.
The first step, \textit{forward process}, transforms data into noise. 
Specifically, the score-based diffusion model~\cite{song2020score} uses the following stochastic differential equations (SDEs)~\citep{sarkka2019applied} for data perturbation:
\begin{align} \label{SDE}
    d\mathbf{x} = f(\mathbf{x}, t) dt + g(t) dW_{t},
\end{align}
where $f(\mathbf{x},t)$ and $g(t)$ are drift and diffusion coefficient, respectively, and 
$W_{t}$ is a standard Wiener-process (a.k.a. Brownian Motion) indexed by time $t\in[0,T]$. 
Here, the $f$ and $g$ functions are user-specified, and~\cite{song2020score} suggests three different types of SDEs, i.e., variance exploding (VE), variance preserving (VP), and sub-variance preserving (sub-VP) SDEs for data perturbation.
Allowing diffusion to continue long enough with $T$ being sufficiently large, it can be shown the distribution of $\mathbf{x}_{t}$ converges to some easy-to-sample distributions $\pi$ such as normal or uniform distributions.
Specifically, when $f:=-\mathbf{x}_{t}$ and $g:=\sqrt{2}$,~\eqref{SDE} is known as Ornstein-Uhlenbeck (OU) process, it is proven $p_{t}:=\text{Law}(\mathbf{x}_{t})\rightarrow \pi$ with $\pi$ being normal, exponentially fast in $2$-Wasserstein distance.
See for instance~\cite{bakry2014analysis}.
However, despite this convergence result, it is analytically difficult to know the exact form of $p_{T}$, and it is often replaced by the $\pi$ in practice when starting the reverse process.
\\ \\
The second step, \textit{reverse process}, is a generative process that reverses the effect of the forward process. 
This process learns to transform the noise back into the data by reversing SDEs in~\eqref{SDE}.
Through the Fokker-Planck equation of marginal density $p_{t}(\mathbf{x})$ for time $t\in[t_{0},T]$, the following reverse SDE~\citep{anderson1982reverse} can be easily derived:
\begin{align}\label{Reverse_SDE}
    d\mathbf{x} = \bigg[ f(\mathbf{x},t) - g(t)^{2} \nabla_{\mathbf{x}} \log p_{t}(\mathbf{x}) \bigg] dt + g(t) d\bar{W}_{t}.
\end{align}
Here, the gradient of $\log p_{t}(\mathbf{x})$ w.r.t to the perturbed data $\mathbf{x}(t)$ is referred to as score function, $dt$ in~\eqref{Reverse_SDE} is an infinitesimal negative time step, and $d\bar{W}_{t}$ is a Wiener-Process running backward in time, i.e., $t:T\rightarrow t_{0}$. 
In practice, $t_{0}$ is usually chosen to be a small number close to $0$, but not too close to $0$ to prevent the blow-up of the score function.
There are various ways to solve~\eqref{Reverse_SDE}; for instance, the discretization scheme such as \textit{Euler-Maruyama}, or the theory-driven method such as \textit{probability-flow}. 
See~\cite{song2020score} for more detailed expositions on these methods. 
The papers introduced in this paper focus on the discretization scheme, and readers can refer~\cite{chen2023probability} for the recent theoretical understandings of the probability flow in the diffusion model. 
\\ \\
The score function, $\nabla_{\mathbf{x}} \log p_{t}(\mathbf{x})$, is approximated by a time-dependent score-based model $\mathbf{S}_{\theta}(\mathbf{x}(t),t)$ which is parametrized by neural networks in practice. 
The network parameter $\theta$ is estimated by minimizing the following score-matching loss:
\begin{align} \label{sml}
    \theta^{\star}:=
    \argmin_{\theta}\mathbb{E}_{t\sim\mathcal{U}[t_{0},T]}\mathbb{E}_{\mathbf{x(t)\mid\mathbf{x(0)}}}\mathbb{E}_{\mathbf{x}(0)}\bigg[\lambda(t)^{2}\left\| \mathbf{S}_{\theta}(\mathbf{x}(t),t)-\nabla_{\mathbf{x}}\log p_{t}(\mathbf{x}(t)\mid\mathbf{x}(0)) \right\|_{2}^{2} \bigg],
\end{align}
where $\mathcal{U}[t_{0},T]$ is a uniform distribution over $[t_{0},T]$, and $\lambda(t)(> 0)$ is a positive weighting function that helps the scales of matching-losses $\left\| \mathbf{S}_{\theta}(\mathbf{x}(t),t) -\nabla_{\mathbf{x}}\log p_{0t}(\mathbf{x}(t)\mid\mathbf{x}(0)) \right\|_{2}^{2}$ to be in the same order across over the time $t\in[t_{0},T]$.
The transition density $p_{t}(\mathbf{x}(t)\mid\mathbf{x}(0))$ is a tractable Gaussian distribution, and $\mathbf{x}(t)$ can be obtained through ancestral sampling~\citep{ho2020denoising}.
\\ \\
Under this setting, we introduce readers to two lines of attempts aimed at finding answers to the following theoretical questions : 
\begin{enumerate}
    \item[Q1.] Can the diffusion model estimate the target distribution $\nu$ via the learned score function? If so, under which conditions on $\nu$, do we have the polynomial convergence guarantees in generalization error bound $\varepsilon$ measured in TV or Wasserstein distances?
    
    \item[Q2.] Do neural networks well approximate and learn the score functions? If so, how one should choose network architectures and what is the sample complexity of learning? Furthermore, if the data distribution has a special geometric structure, is the diffusion model adaptive to the structure, just as GAN models do? 
\end{enumerate}
The main statistical object of interest in these two lines of research is the generalization bound measuring the distance between target distribution $\nu$ and estimated distribution $\widehat{\mu}_{\theta}$ from the samples $\mathbf{x}_{t_{0}}$ by solving the reverse SDE in~\eqref{Reverse_SDE}.
Here, the score function is substituted by the estimated time-dependent neural network $\mathbf{S}_{\theta}(\mathbf{x}(t),t)$. 
The first line of work mainly focuses on the sampling perspective of the diffusion model, given that we have good estimates of the score function.  
The second line of work extends the attention to the score function approximation through neural networks.
Furthermore, under some highly stylized settings, they specify the explicit network structures which give good generalization guarantees. 
\\ \\
\noindent \textit{\textbf{{Attempts to answer Q1. }}}
Early theoretical efforts to understand the sampling of score-based diffusion model suffered from being either not quantitative~\citep{de2021diffusion, liu2022let}, or from 
the curse of dimensionality~\citep{de2022convergence, block2020generative}.
Specifically among them,~\cite{de2022convergence} gave the convergence in $1$-Wasserstein distance for distributions with bounded support $\mathcal{M}$.
Their case covers the distributions supported on lower-dimensional manifolds, where guarantees in TV or KL distance are unattainable as there are no guarantees that $\nu$ and $\widehat{\mu}_{\theta}$ have the same support set.
For general distributions, their bound on $W_{1}(\nu,\widehat{\mu}_{\theta})$ have the exponential dependence on the diameter of the manifold $\mathcal{M}$ and truncation of reverse process $t_{0}$ as $\mathcal{O}(\exp(\text{diam}(\mathcal{M})^{2}/t_{0}))$.
For smooth distributions where the Hessian $\nabla^{2}\log p_{t}$ is available, the bound is further improved with a polynomial dependence on $t_{0}$ with the growth rate of Hessian as $t\rightarrow 0$ being on the exponent.
\\ \\
To the best of our knowledge,~\cite{lee2022convergence} first gives the polynomial guarantees in TV distance under $L^{2}$-accurate score for a reasonable family of distributions. 
However, their result is based on the assumption that the distribution meets certain smoothness criteria and the log-Sobolev inequality, which essentially confines the applicability of their findings to distributions with a single peak.
Recently, two works~\cite{lee2023convergence, chen2022sampling} appear online trying to avoid the strong assumptions on the data distributions and to get the polynomial convergence guarantees under general metrics such as TV or Wasserstein. 
Specifically,~\cite{lee2023convergence} give $2$-Wasserstein bounds for \textit{any} distributions with bounded support.
Contrary to~\cite{de2022convergence,lee2022convergence}, the results they provide have polynomial complexity guarantees without relying on the functional inequality on distributions such as log-Sobolev inequality.
They further give TV bounds with polynomial complexity guarantees under the Hessian availability assumption.
Similarly with~\cite{lee2022convergence}, under general data distribution assumption, i.e., second moment bound of $\nu$ and $L$-Lipschtizness of score function,~\cite{chen2022sampling} give the polynomial TV convergence guarantee, where only $\Tilde{\Theta}\big( \frac{L^{2}d}{\varepsilon^{2}} \big)$ discretization is needed.
Here, $\varepsilon$ is a TV generalization error, $d$ being a data dimension. 
\\ \\
\noindent \textit{\textbf{{Attempts to answer Q2. }}}
Due to its recent theoretical advancements, the list of research attacking the second question is short in the literature. 
One work~\citep{chen2023score} proved that the diffusion model is adaptive to estimating the data distribution supported in a lower-dimensional subspace.
They design a very specific network architecture for $\mathbf{S}_{\theta}(\mathbf{x}(t),t)$ with an encoder-decoder structure and a skip-connection.  
Under a more general setting, another work~\citep{oko2023diffusion} proves the distribution estimator from the diffusion model can achieve nearly minimax optimal estimation rates. 
Specifically, they assume the true density is supported on $[-1,1]^{d}$, in the Besov space with a smooth boundary. 
The Besov space unifies many general function spaces such as H\"older, Sobolev, continuous, or even non-continuous function classes. (Also refer section~\ref{Approx_guarantees}.)
The result in~\cite{oko2023diffusion} is valid for the non-continuous function class, and this should be contrasted with the aforementioned works~\citep{lee2023convergence,chen2022sampling} where they assume the Lipschitzness of the score function.
The exact architecture on the score network is also given in the form of~\eqref{network_def}.

\subsection{In-Context Learning in Large Language Model}
We provide readers with recent theoretical understandings of an interesting phenomenon observed in LLM, i.e., 
\textit{In-Context Learning} (ICL). It refers to the ability of LLMs conditioned on prompt sequence consisting of examples from a task (input-output pairs) along with the new query input, the LLM can generate the corresponding output accurately. 
An instance taken from~\cite{garg2022can}, these models can produce English translations of French words after being prompted on a few such translations, e.g.:
\begin{align*}
    \underbrace{
    \text{maison}\rightarrow \text{house}, \quad
    \text{chat}\rightarrow \text{cat}, \quad
    \text{chien}\rightarrow}_{\text{prompt}} 
    \underbrace{\text{dog}}_{\text{completion}}.
\end{align*}
This capability is quite intriguing as it allows models to adapt to a wide range of downstream tasks on-the-fly without the need to update the model weights after training.
Readers can refer to the backbone architecture of LLM  (i.e., Transformer) in the seminal paper~\citep{vaswani2017attention}.
\\ \\
Toward further understanding ICL, researchers formulated a well-defined problem of learning a function class $\mathcal{F}$ from in-context examples.
Formally, let $\mathcal{D}_{\mathcal{X}}$ be a distribution over inputs and $\mathcal{D}_{\mathcal{F}}$ be a distribution over functions in $\mathcal{F}$.
A prompt $P$ is a sequence $(x_{1}, f(x_{1}), \dots, x_{k}, f(x_{k}), x_{\text{query}} )$ where inputs (i.e., $x_{i}$ and $x_{\text{query}}$) are drawn i.i.d. from $\mathcal{D}_{\mathcal{X}}$ and $f$ is drawn from $\mathcal{D}_{\mathcal{F}}$.
In the above example, it can be understood $\{(x_{1},f(x_{1}),(x_{2}, f(x_{2})\}:=\text{\{(maison, house), (chat, cat)\}}$, $x_{\text{query}}=\text{chien}$, and $f(x_{\text{query}})=\text{dog}$.
Now, we provide the formal definition of in-context learning. 

\begin{definition} [In-context learning~\cite{garg2022can}]
    Model $M_{\theta}$ can in-context learn the function class $\mathcal{F}$ up to $\varepsilon$, with respect to $(\mathcal{D}_{\mathcal{F}},\mathcal{D}_{\mathcal{X}})$, if it can predict $f(x_{\text{query}})$ with an average error
    \begin{align} \label{ICL}
        \mathbb{E}_{P\sim(x_{1}, f(x_{1}), \dots, x_{k}, f(x_{k}), x_{\text{query}} )}\big[ \ell(M_{\theta}(P),f(x_{\text{query}})) \big] \leq \varepsilon, 
    \end{align}
    where $\ell(\cdot, \cdot)$ is some appropriate loss function, such as the squared error.
\end{definition}
\noindent
The paper~\cite{garg2022can} empirically investigated the ICL of Transformer architecture~\citep{vaswani2017attention} by training the model $M_{\theta}$ on random instances from linear functions, two-layer ReLU networks, and decision trees. 
Here, we omit the definition of Transformer for brevity. 
Specifically, they showed the predictions of Transformers on the prompt $P$ behave similarly to those of ordinary least-squares (OLS), when the models are trained on the instances from linear function classes $\mathcal{F}^{\text{Lin}}:=\{f\mid f(x)=w^{\top}x,  w\in\mathbb{R}^{d}\}$ for random weights $w\sim\mathcal{N}(0, I_{d})$. 
A similar phenomenon was observed for the models trained on sparse linear functions as the predictions behave like those of Lasso estimators. 
\\ \\
These nice observations sparked numerous follow-up theoretical studies of ICL on internal mechanisms~\citep{akyurek2022learning, von2023transformers,dai2022can}, expressive power~\citep{akyurek2022learning,giannou2023looped} and generalizations~\cite{li2023transformers}.
Among them, papers~\citep{akyurek2022learning, von2023transformers} investigated the behavior of transformers when trained on random instances from $\mathcal{F}^{\text{Lin}}$, and showed the trained Transformer's predictions mimic those of the single step of gradient descent. 
They further provided the construction of Transformers which implements such a single step of gradient descent update.
The recent paper~\cite{zhang2023trained} explicitly proved the model parameters estimated via gradient flow converge to the global minimizer of the non-convex landscape of population risk in~\eqref{ICL} for learning $\mathcal{F}^{\text{Lin}}$.
Nonetheless, the results in the paper are based on a linear self-attention (LSA) layer without softmax non-linearities and simplified parameterizations.
Later,~\cite{huang2023context} considered the single head attention with softmax non-linearity and proved the trained model through gradient descent indeed in-context learn $\mathcal{F}^{\text{Lin}}$ under highly stylized scenarios (i.e., simplified parameter settings, orthonormal features).
Recently,~\cite{chen2024training} generalized the setting to the multi-head attention layer with softmax non-linearity for in-context learning multi-task linear regression problems.
\\ \\
One important work~\cite{bai2023transformers} showed the existence of Transformers that can implement a broad class of standard machine learning algorithms in-context such as least squares, ridge regression, Lasso, and gradient descent for two-layer neural networks. 
This paper goes beyond and demonstrates a remarkable capability of a single transformer: the ability to dynamically choose different base in-context learning (ICL) algorithms for different ICL instances, all without requiring explicit guidance on the correct algorithm to use in the input sequence. This observation is noteworthy as it mirrors the way statisticians select the learning algorithms for inferences on model parameters. 

\section{Conclusions \&  Future Topics} \label{Conc}
In this article, we review the papers that studied neural networks mainly from statistical viewpoints. 
In Section~\ref{Approx_theory}, we review statistical literature that primarily relies on approximation-theoretic results of neural networks. This framework allows for interesting comparisons between neural networks and classical linear estimators in various function estimation settings. 
Specifically, neural networks are highly adaptive to functions with special geometric structures, whereas classical linear estimators are not. 
See Figure~\ref{Fig0}.
In Section~\ref{training_dynamics}, we review papers studying the statistical guarantees of neural networks trained with gradient-based algorithms. The overparameterization of neural networks impacts the landscape of loss function, streamlining the mathematical analysis of training dynamics. We discuss training dynamics in two regimes: Neural Tangent Kernel (NTK) and Mean-Field (MF). Specifically, some works which studied how networks in NTK regime can offer statistical guarantees under noisy observations are introduced. We also introduce attempts to unify and go beyond these regimes, explaining the success of networks with finite widths.
In Section~\ref{Generative_Models}, we review the statistical guarantees of deep generative models (i.e., GAN and diffusion model) for estimating the target distributions. 
Neural networks form the fundamental backbone of both frameworks, enabling the adaptive estimation of distributions with specialized structures.
Some statistical works on ICL phenomena observed in LLM are also introduced, 
\\ \\
\noindent However, aside from these topics, several promising avenues have not been covered in this paper.
We will briefly them for the references of readers.
\\ \\
\noindent\textit{\textbf{{Generative Data Science.}}} 
In modern machine learning (ML), data is valuable but often scarce and has been referred to as \textit{"the new oil"}, a metaphor credited to mathematician Clive Humby. 
With the rise of deep generative models like GANs, diffusion models, and LLMs, synthetic data is rapidly filling the void of real data and finding extensive utility in various downstream applications.
For instance, in the medical field, synthetic data has been utilized to improve patients' data privacy and the performance of predictive models for disease diagnosis~\citep{chen2021synthetic}. 
Similarly, structured tabular data is the most common data modality that requires the employment of synthetic data to resolve privacy issues~\citep{li2023statistical,suh2023autodiff} or missing values~\citep{ouyang2023missdiff}.
Furthermore, Synthetic data has been at the core of building reliable AI systems, specifically for the promising issues in fairness \citep{zeng2022bayes}, and robustness \citep{ouyang2023improving}. 
Despite its prevalence in the real world, how to evaluate synthetic data from the dimensions of fidelity, utility and privacy-preserving remains unclear. Specifically, we want to address the following general questions: 
\begin{enumerate}
    \item[Q1] How well the models trained via synthetic data generalize well to real unseen data? (e.g.,~\cite{xu2023utility})?
    \item[Q2] How does artificially generated data perform in the various downstream tasks such as classification or regression? 
    (e.g.,~\cite{li2023statistical, shirong2023binary}), and adversarial training (e.g.,~\cite{xing2022artificially,xing2022unlabeled})?

    \item[Q3] How does the synthetic data generated to satisfy certain privacy constraints (e.g., differential privacy~\cite{dwork2008differential})? 
\end{enumerate}

Addressing these questions systemmatically requires establishing a new framework of ``Generative Data Science", aiming to elucidates the underlying principles behind generative AI. As evidenced by the above referenced works, this vision is supported by the recent observation that ``creating something out of nothing" is possible and beneficial through synthetic data generation.
\\ \\
\noindent\textit{\textbf{{Kolmogorov Arnold Network (KAN).}}} 
As of June 2024, a new type of architecture, KAN~\citep{liu2024kan}, appears on arXiv and has been receiving enormous attention from the ML community.
The model is motivated by the Kolmogorov-Arnold Representation Theorem (KART), stating any continuous and smooth functions on the bounded domain can be represented as compositions and summations of the finite number of univariate functions. 
Several papers have explored the connections between neural networks and the KART due to their similarities in terms of compositional structure.
For instance, see~\cite{poggio1989theory,girosi1989representation,schmidt2021kolmogorov} and references therein.
The authors claim that KAN outperforms fully-connected networks in terms of accuracy and interpretability for function approximations on their specially designed tasks.
Nonetheless, this model definitely requires further research for better use in the future for both practical and theoretical purposes.
From a practical perspective, KAN's training time is 10 times slower than fully-connected networks, and authors only apply the model to small-size tasks. 
Interested readers can refer to Section $6$ of~\cite{liu2024kan}.
From a theoretical point of view, the authors claim that KAN avoids the curse of dimensionality for function approximation, whereas fully-connected networks cannot. 
But this argument requires further investigation under more rigorous settings with various types of function spaces $\mathcal{G}$.

\section*{DISCLOSURE STATEMENT}
The authors are not aware of any affiliations, memberships, funding, or financial holdings that
might be perceived as affecting the objectivity of this review. 

\section*{ACKNOWLEDGMENTS}
We would like to thank an anonymous reviewer and Minshuo Chen for insightful comments on the draft of this review.
This survey is partially sponsored by NSF – SCALE MoDL (2134209), NSF – CNS (2247795), Office of Naval Research (ONR N00014-22-1-2680) and CISCO Research Grant.

\begin{appendices}
\begin{centering} \section*{\LARGE Appendix} \end{centering}
\noindent We defer some of the technical contents included in the first version of this draft to the Appendix.
Following is the list of materials to be covered 
\\
\begin{enumerate}
    \item[1.] We introduce the literature which studied the statistical guarantees of neural networks based on approximation-theory, specifically a result of~\cite{schmidt2020nonparametric} in the Appendix~\ref{regression}. 

    \item[2. ] Additional survey on statistical research of approximation-theory based classification tasks is presented in the Appendix~\ref{classification}. 
    
    \item[3.] The intuition of PDE (Equation.~\eqref{PDE}) in the Mean-Field Regime is explained through a rough derivation process in the Appendix~\ref{PDE_intuition}.

    \item[4.] More detailed descriptions of the problem settings and results of some works introduced in Subsection 3.3. in the main manuscript are provided in the Appendices~\ref{AppendixC} and \ref{AppendixD}.
\end{enumerate}

\section{Technical Results of~\cite{schmidt2020nonparametric}} \label{regression}
In this section, we would like to provide readers with a little bit of detailed descriptions on the seminal paper~\cite{schmidt2020nonparametric}. 
The paper assumed that the regression function $f_{\rho}$ has a special structure; a hierarchical compositional structure:
\begin{eqnarray} \nonumber
    f_{\rho} = g_q \circ g_{q-1} \circ g_{q-2} \circ \dots \circ g_{0},
\end{eqnarray}
with $g_i:[a_i,b_i]^{d_i}\rightarrow{[a_{i+1},b_{i+1}]^{d_{i+1}}}$. Denote by $g_i=(g_{ij})^{\text{T}}_{j=1,\dots,d_{i+1}}$ the components of $g_i$ and let $t_i$ be the maximal number of variables on which each of the $g_{ij}$ depends on.
This setting leads to $t_{i}\leq d_{i}$. 
Each of the $d_{i+1}$ components of $g_{i}$ belongs to  $r_{i}$-H\"{o}lder class. 
Finally, the underlying function space to be considered is 
\begin{align*} 
    \mathcal{G}(q,\mathbf{d},\mathbf{t},\mathbf{r},K):=\big\{f_{\rho} = g_q &\circ g_{q-1} \circ \dots \circ g_{0} : g_{i}=(g_{ij})_{j}:[a_i,b_i]^{d_i} \rightarrow{[a_{i+1},b_{i+1}]^{d_{i+1}}},\\ &g_{ij}\in\mathcal{H}^{r_{i}}([a_i,b_i]^{t_i},K), \quad \forall |a_i|,|b_i|\leq K \big\}, 
\end{align*}
where $\mathbf{d}:=(d_{0},\dots,d_{q+1})$, $\mathbf{t}:=(t_{0},\dots,t_{q})$, and $\mathbf{r}:=(r_{0},\dots,r_{q})$.
By~\cite{juditsky2009nonparametric,ray2017regularity}, importantly, the induced-smoothness on $f_{\rho}$ is driven as follows: 
\begin{align}
    r_{i}^{\star}:=r_{i}\prod_{\ell=i+1}^{q}\big(r_{\ell} \wedge 1 \big), \quad \forall i\in\{0,\dots,q\}.
\end{align}

In Theorem $3$ of \cite{schmidt2020nonparametric}, it is proven that the rate $\phi_{n}:=\max_{i=0,\dots,q}n^{-\frac{2 r^{\star}_{i}}{2 r^{\star}_{i}+t_{i}}}$ is the minimax optimal rate over the class $\mathcal{G}(q,\mathbf{d},\mathbf{t},\mathbf{r},K)$ in the interesting regime $t_{i}\leq \text{min}\big( d_{0},\dots,d_{i-1} \big)$ for all $i$.
This regime avoids the $t_{i}$'s being larger than the input dimension $d_{0}$.
Under this setting on the regression function, the author proves that there exists an empirical risk minimizer $\widehat{f}_{n}\in\mathcal{F}(L,\mathbf{p}, \mathcal{N})$ with $L\asymp \log n$, $\mathbf{p}_{\text{max}}\asymp n^{C}$ with $C \geq 1$ and $\mathcal{N}\asymp n\phi_{n}\log n$ such that it achieves the nearly minimax optimal rate, i.e., $C^{\prime}\phi_{n}L\log^{2}n$ for the convergence on the excess risk. 
The constant factor $C^{\prime}$ depends on $q,\mathbf{d},\mathbf{t},\mathbf{r},K$.\
Importantly, they proved a classic statistical method such as Wavelet cannot achieve the minimax rate, showing the superior adaptiveness of neural networks for estimating $f_{\rho}\in\mathcal{G}(q,\mathbf{d},\mathbf{t},\mathbf{r},K)$.
\\ \\
Several remarks follow in the sequel. 
First, the rate $\phi_{n}$ is dependent on the effective dimension $t_{i}$ which can be much less than the ambient input dimension $d_{0}$, and this implies that the neural net can circumvent the curse of dimensionality.
Second, the depth $L$ should be chosen to scale with the sample size in a rate $\mathcal{O}(\log n)$.
Third, the width $\mathbf{p}_{\text{max}}$ can be chosen to be independent of the smoothness indices.
Lastly, the result is saying what is important for the statistical performance is not the sample size, but the amount of regularization.
This is explicitly reflected through the number of active parameters $\mathcal{N}\asymp n\phi_{n}\log n \ll n$.

\section{Classification tasks via fully-connected networks} \label{classification}
We consider a binary classification problem via fully-connected networks. Classifiers built with neural networks handle large-scale high-dimensional data, such as facial images from computer vision extremely well, while traditional statistical methods often fail miserably.
We will review papers that have attempted to provide theoretical explanations for such empirical successes in high-dimensional classification problems, beyond the existing statistical literature. This is a less studied area than nonparametric regression, although deep learning is often used for classification. 
\\ \\
\noindent \textit{\textbf{{Binary Classification Task.} }}
Let $\{X_i,Y_i\}_{i=1}^{n}$ be i.i.d. random pairs of observations, where $X_{i} \in \mathbb{R}^{d}$ and $Y_{i}\in \{0,1\}$.
Denote $P_X$ the probability distribution of $X_i$ and $\pi=\pi_{X,Y}$ the joint distribution of $(X,Y)$.
In a classification problem, it is of main interest to estimate the decision rule, where it is determined through a set $G$.
In this paper, we assume $G$ is a Borel subset of $\mathbb{R}^{d}$, where it gives $Y=1$ if $X\in G$ and $Y=0$ if $X\notin G$. 
The misclassification risk associated with $G$ is given by: 
\begin{eqnarray*}
    R(G) := P(Y\neq \mathbbm{1}(X\in G)) = \mathbb{E}[(Y-\mathbbm{1}(X\in G))^{2}].
\end{eqnarray*}
It is widely known that a Bayes classifier minimizes $R(G)$ denoted as $G^{*}_{\pi}=\{x:\eta^{\star}(x)\geq 1/2\}$, where $\eta^{\star}(x)=P(Y=1|X=x)$.
However, since we have no information on the joint distribution $\pi$, it is difficult to directly find the minimizer of $R(G)$.
There are two ways to circumvent this difficulty.
Given the data set $\{X_i,Y_i\}_{i=1}^{n}$, the first method is to directly estimate the set minimizing the empirical risk;
\begin{eqnarray*}
    R_n(G) = \frac{1}{n}\sum_{i=1}^{n}\big(Y_{i}-\mathbbm{1}(X_{i}\in G)\big)^{2}.
\end{eqnarray*}
We denote the minimizer of $R_n(G)$ as $\widehat{G}_{n}$.
The second method is to estimate the {regression function}, $\widehat{\eta}(x)$, and then plug in the $\widehat{\eta}(x)$ to the set $G$ denoted as $\widehat{G}_{n}:=\{x\in \mathbb{R}^{d} : \widehat{\eta}(x) \geq 1/2\}$.
Under these settings, we are interested in how fast $R(\widehat{G}_{n})$ converges to $R(G_\pi^{*})$ as $n\rightarrow{\infty}$. 
Note that these two approaches are different in the sense that they impose different assumptions to derive the convergence rate of $R(\widehat{G}_{n})$ to $R(G_\pi^{*})$.

Now, we state three assumptions commonly imposed on the joint distribution $\pi$.
\begin{enumerate}    
    \item $\emph{\textbf{C}omplexity \textbf{A}ssumption on the \textbf{D}ecision set}$ \textbf{(CAD)} characterizes the smoothness of boundary of Bayes classifier $G_{\pi}^{\star}(\subset \mathcal{G})$ by assuming the class $\mathcal{G}$ has a suitably bounded $\epsilon$-entropy.
    
    \item $\emph{\textbf{C}omplexity \textbf{A}ssumption on the \textbf{R}egression function}$ \textbf{(CAR)} refers the regression function $\eta^{\star}(x)$ is smooth enough and  belongs to a function class $\Sigma$ having a suitably bounded $\epsilon$-entropy.
    
    \item $\emph{\textbf{M}argin \textbf{A}ssumption}$ (\textbf{MA}) describes the behavior of $\eta^{\star}(x)$ near the decision boundary (i.e., $\eta^{\star}(x)=1/2$).
    Specifically, the assumption parameter $q(\geq 0)$ describes how much the regression is bounded away from $1/2$. 
    The larger the $q$ is, the farther $\eta^{\star}(x)$ is bounded away from $1/2$.
\end{enumerate}

For those who are interested in rigorous treatments of each assumption, we refer readers~\cite{mammen1999smooth,tsybakov2004optimal,audibert2007fast} and references therein. 
Note that, in general, there is no connection between assumptions (CAR) and (CAD). 
Indeed, the fact that $G^*$ has a smooth boundary does not imply that $\eta^{\star}(x)$ is smooth, and vice versa.
However, as noted by~\cite{audibert2007fast}, the smoothness of $\eta^{\star}(x)$ and (MA) parameter $q$ cannot be simultaneously large; as the smooth $\eta^{\star}(x)$ hits the level $1/2$, it cannot "take off" from this level too abruptly.
\\ \\
\noindent \textit{\textbf{{Results on Neural Networks.}}}
The Table 1 summarizes the settings and results of the works that study the convergence rates of excess risks for binary classification problems of neural networks. 
Hereafter, $\alpha>0$ indicates the smoothness index of either target function class $(\Sigma)$ in CAR or boundary class $(\mathcal{G})$ in CAD, and $q>0$ denotes the term in the exponent in Margin Assumption (MA).

\begin{table}[!t] \label{Table_classification}
\begin{adjustbox}{width=\linewidth, center}
\begin{tabular}{|c|c|c|c|c|} 
\hline
Reference         & Function Space             & Loss                   & Condition         & Rate \\ \hline
\multirow{2}{*}{\cite{kim2021fast}} & \multirow{2}{*}{ReLU FNNs} & \multirow{2}{*}{Hinge} &  \textbf{MA, CAD}                                                    &  $\mathcal{O}\bigg( n^{-\frac{\alpha(q+1)}{\alpha(q+2)+(d-1)(q+1)}}\bigg)$     \\ \cline{4-5} 
                  &                            &                        & \textbf{MA, CAR}                                          &  $\mathcal{O}\bigg( n^{-\frac{\alpha(q+1)}{\alpha(q+2)+d}}\bigg)$     \\ \hline
\multirow{3}{*}{\cite{feng2021generalization}} & \multirow{6}{*}{ReLU CNNs} & Hinge                  & \multirow{2}{*}{\textbf{CAR}} &   $\mathcal{O}\bigg( n^{-\frac{r}{2\beta(d-1)+r(2-\tau)}}\bigg)$   \\  \cline{3-3} \cline{5-5} 
                  &                            & p-norm                 &                                                                                                                                                       &  $\mathcal{O}\bigg( n^{-\frac{pr}{2(\beta+1)(d-1)+2pr(2-\tau)}}\bigg)$     \\  \cline{3-5} 
                  &                            & 2-norm                 & \textbf{MA, CAR}                                              &  $\mathcal{O}\bigg( n^{-\frac{2rq}{(2+q)((\beta+1)(d-1)+2\tau)}}\bigg)$     \\ \cline{1-1} \cline{3-5} 
\multirow{3}{*}{\cite{shen2022approximation}} &                            & Hinge                  & \textbf{MA, CAR}                                                      &  $\mathcal{O}\bigg( n^{-\frac{r(q+1)}{d+2r(q+1)}}\bigg)$     \\  \cline{3-5} 
                  &                            & Logistic               & \multirow{2}{*}{\textbf{CAR}}                                                                                                  &  $\mathcal{O}\bigg( n^{-\frac{r}{2d+4r}}\bigg)$     \\  \cline{3-3} \cline{5-5} 
                  &                            & Least Square           &                                                                                                                                                       &  $\mathcal{O}\bigg( n^{-\frac{4r}{3d+16r}}\bigg)$     \\ \hline
\cite{zhou2023classification}  & ReLU FNNs                  & Hinge                  & \begin{tabular}[c]{@{}c@{}}\textbf{MA}: \\ \text{GMM} \end{tabular}                                                                                     &  $\mathcal{O}\bigg( n^{-\frac{q+1}{q+2}}\bigg)$     \\ \hline
\cite{ko2023classification}    & ReLU FNNs                  & Logistic                  & \begin{tabular}[c]{@{}c@{}}\textbf{MA, CAR}: \\ \text{BA} \end{tabular}                                                                                     &  $\mathcal{O}\bigg( n^{-\frac{q+1}{3(q+2)}}\bigg)$     \\ \hline
\cite{hu2020optimal}    & ReLU FNNs                  & 0-1 Loss         & \begin{tabular}[c]{@{}c@{}}\textbf{MA, CAD}: \\ \text{Teacher-Student} \end{tabular}                                                                                     &  $\mathcal{O}\bigg( n^{-\frac{2}{3}}\bigg)$     \\ \hline
\end{tabular}
\end{adjustbox}
\caption{Comparison table on statistical convergence rates of neural networks in classification tasks.}
\end{table}

The first notable paper~\cite{kim2021fast} proved that under the smooth boundary (CAD) and smooth regression (CAR) conditions, respectively, there exist neural networks that achieve (nearly) optimal convergence rates of excess risks.
The obtained rate under CAR assumption is sub-optimal in a sense~\cite{tsybakov2004optimal} as they prove the minimax optimal rate achievable by neural nets is $\mathcal{O}\big( n^{-\alpha(q+1)/[\alpha(q+2)+(d-1)q]}\big)$, and the optimal rate under CAD assumption is achievable.
According to their remark, no other estimators achieve fast convergence rates under these scenarios simultaneously. 
\\ \\
Two recent works studied the binary classification by ReLU convolutional neural networks~\cite{feng2021generalization,shen2022approximation}.
The paper~\cite{feng2021generalization} considered the $p$-norm loss $\phi(t):=\max\{0,1-t\}^{p}$ for $p\geq 1$, and the input data is supported on the sphere $\mathcal{S}^{d-1}$.
In the paper, the approximation error bounds and excess risk bounds are derived under a varying power condition, and the target function class is the Sobolev space $W_{p}^{r}(\mathcal{S}^{d-1})$ for $r>0$ and $p\geq 1$.
Technically, they obtained the approximation error in $L^{p}$ norm including the case $p=\infty$.
Two quantities including $\beta =\max\{1,(d+3+r)/(2(d-1))\}$ and $\tau\in[0,1]$ are involved in the rate.
The paper~\cite{shen2022approximation} established the convergence rates of the excess risk for classification with a class of convex loss functions.
The target function of interests is also from the Sobolev space $W_{p}^{r}([0,1]^{d})$, and note that the feature domain is $[0,1]^{d}$. 
Under this setting, the most interesting aspect of this work is that they track the dependencies of the ambient dimension $d$ in the pre-factor hidden in the big-$\mathcal{O}$ notation.
They showed the prefactor is polynomially dependent on $d$.
\\ \\
The last three works in the above table give the dimension-free rates in the exponent of $n$ under different problem settings.
The paper~\cite{zhou2023classification} studied the problem of learning binary classifiers through ReLU FNNs. An interesting aspect of this result is that unlike the other results~\cite{kim2021fast,feng2021generalization,shen2022approximation}, the rate is both dimension and smoothness index free, despite the feature domain being unbounded.
The main idea for getting this result is to leverage the fact that Gaussian distributions of features are analytic functions and have a fast decay rate, and this can be captured through ReLU FNNs. Another paper~\cite{ko2023classification} works on a similar problem but under a different setting. 
Notably, they are interested in the Barron Approximation (BA) space proposed in~\cite{caragea2023neural}.
Unlike the classical Barron space in~\cite{barron1993universal}, which is essentially a subset of a set of Lipschitz continuous functions, this space includes even discontinuous functions and hence is more general. 
The CAR assumption is imposed on the functions in the BA class. 
They proved that the rate $\mathcal{O}(n^{-\frac{q+1}{3(q+2)}})$ is achievable through ReLU FNNs for the estimation, and this rate is indeed minimax optimal. 
Note that the rate is slower than the parameteric rate $n^{-\frac{1}{2}}$ by noting the rate ranges from $n^{-\frac{1}{6}}$ to $n^{-\frac{1}{3}}$ as $q$ varies from $0$ to $\infty$.
This is attributed to the setting that the size of the distribution class is large. 
\\ \\
Lastly, a paper~\cite{hu2020optimal} studied the convergence rate of excess risk of neural networks under the framework of~\cite{mammen1999smooth}. 
Specifically, the authors of the paper consider a Teacher-Student framework. 
In this setup, one neural network, called student net, is trained on data generated by another neural network, called teacher net.
Adopting this framework can facilitate the understanding of how deep neural networks work as it provides an explicit target function with bounded complexity. 
Furthermore, assuming the target classifier to be a teacher network of an explicit architecture might provide insights into what specific architecture of the student classifier is needed to achieve an optimal excess risk.
Under this setting, the rate of convergence is derived as 
$\Tilde{\mathcal{O}}_{d}(n^{-\frac{2}{3}})$ for the excess risk of the empirical $0-1$ loss minimizer, given that the student network is deeper and larger than the teacher network.
The authors use the notation $\Tilde{\mathcal{O}}_{d}$ to denote the rate is dependent on input dimension $d$ in a logarithmic factor.
When data are separable, the rate improves to $\Tilde{\mathcal{O}}_{d}(n^{-1})$. In contrast, as it has already been shown in~\cite{mammen1999smooth}, under CAD assumption, the optimal rate is $\mathcal{O}(n^{-\alpha(q+1)/[\alpha(q+2)+(d-1)q]})$.
Clearly, this rate suffers from ``Curse of Dimensionality'' but interestingly, coincides with the rate $\Tilde{\mathcal{O}}_{d}(n^{-\frac{2}{3}})$ when $\alpha=1$ and $\beta\rightarrow{\infty}$.
If we further allow $\alpha\rightarrow{\infty}$ (corresponding to separable data), the classical rate above recovers $\mathcal{O}(n^{-1})$. 
The paper also got the rate is un-improvable in a minimax sense by showing that the minimax lower bound has the same order as that of the upper bound (i.e., $\Tilde{\mathcal{O}}_{d}(n^{-\frac{2}{3}})$).

\section{Intuition of PDE~Equation (Distributional Dynamics) in~\cite{mei2018mean}} \label{PDE_intuition}

In the Mean-Field setting, recall the trajectories of empirical distribution of $\theta^{(k)}$ denoted as $\widehat{\rho}^{(M)}_{k}:=\frac{1}{M}\sum_{r=1}^{M}\delta_{\theta^{(k)}_{r}}$ weakly converges to the deterministic limit $\rho_{t}\in\mathcal{P}(\mathbb{R}^{\mathcal{D}})$ as $k\rightarrow{\infty}$ and $M\rightarrow{\infty}$.
The measure $\rho_{t}$ is the solution of the following nonlinear partial differential equation (PDE):
\begin{align} 
   &\partial_{t}\rho_{t}= \nabla_{\mathbf{\theta}}\cdot\big(\rho_{t} \nabla_{\mathbf{\theta}} \mathbf{\Psi}(\mathbf{\theta};\rho_{t})\big),
    \qquad \mathbf{\Psi}(\mathbf{\theta};\rho_{t}):=\mathcal{V}(\mathbf{\theta}) + \int\mathcal{U}(\mathbf{\theta},\Bar{\mathbf{\theta}})\rho_{t}(d\Bar{\mathbf{\theta}}), \label{PDE} \\
    &\qquad \mathcal{V}(\mathbf{\theta}):=-\mathbb{E}\{\mathbf{y}\sigma_{\star}(\mathbf{x};\mathbf{\theta})\},
    \qquad \qquad \mathcal{U}(\mathbf{\theta_{1}},\mathbf{\theta_{2}}):=\mathbb{E}\{\sigma_{\star}(\mathbf{x};\mathbf{\theta}_{1})\sigma_{\star}(\mathbf{x};\mathbf{\theta}_{2})\}. \nonumber
\end{align}
The PDE in~Equation.~\eqref{PDE} is often referred as \textit{Distributional Dynamics} (DD). 
The intuition of the above PDE can be understood through the derivation process.
To aid the understandings of readers, we only present a high-level idea for the derivation without rigor.
For each particle $\theta_{r}$, we can write the $k$th update of one-pass SGD as follows:
\begin{align*}
    \theta_{r}^{k+1} = \theta_{r}^{k} + 2 \eta_{k} \nabla_{\theta_{r}}\sigma_{\star}(\mathbf{x}_{k},\theta_{r}^{k})\bigg\{ \mathbf{y}_{k}-\frac{1}{M}\sum_{r=1}^{M}\sigma_{\star}(\mathbf{x}_{k};\mathbf{\theta}_{r}^{k})\bigg\}.
\end{align*}
By fixing the learning rate $\eta_{k}:=\frac{\varepsilon}{2}$ with $\varepsilon$ close to $0$ and taking expectation with respect to data $(\mathbf{x},\mathbf{y})$ on both sides of the equality, we have the ODE desribing the continuum motion of each particle $\theta_{r}$:
\begin{align} \label{ODE_Mean_Field}
    \dot{\mathbf{\theta}}_{r}(t)=-2\nabla_{\mathbf{\theta_{r}}}\mathcal{V}(\mathbf{\theta_{r}})-\frac{2}{M}\sum_{k=1}^{M}\nabla_{\mathbf{\theta_{r}}}\mathcal{U}(\mathbf{\theta_{r}},\mathbf{\theta_{k}}).
\end{align}
Each particle $(\theta_{r})$ moves to the directions where it minimizes $-\mathcal{V}(\mathbf{\theta_{r}}):=\mathbb{E}\{\mathbf{y}\sigma_{\star}(\mathbf{x};\mathbf{\theta})\}$ to secure the best fit with the response data $\mathbf{y}$.
At the same time, it moves to the directions where it minimizes $-\frac{1}{M}\sum_{k=1}^{M}\mathcal{U}(\mathbf{\theta_{r}},\mathbf{\theta_{k}})$ so that 
the particle $\mathbf{\theta}_{r}$ does not get too close to the other particles $(\mathbf{\theta}_{k})_{k\neq r}$.
Recalling the definition of $\mathbf{\Psi}(\mathbf{\theta};\rho_{t})$, we rewrite the ODE~Equation.~\eqref{ODE_Mean_Field} in terms of empirical distribution $\widehat{\rho}^{(M)}_{k}$:
\begin{align}
    \dot{\mathbf{\theta}}_{r}(t):=-\nabla_{\theta_{r}}\mathbf{\Psi}(\mathbf{\theta}_{r};\widehat{\rho}^{(M)}_{k}).
\end{align}
Then, the authors introduced new $M$ i.i.d. trajectories $(\Bar{\theta}_{r}(t))$ with $r=1,\dots,M$ with same initializations as for the SGD (i.e., $\Bar{\theta}_{r}(0)={\theta}_{r}(0)$), where the trajectory $\Bar{\theta}_{r}(t)$ is the solution of the ODE $\dot{\mathbf{\Bar{\theta}}}(t):=-\nabla_{\Bar{\theta}}\mathbf{\Psi}(\Bar{\theta}(t);\rho_{t})$, then proved $\Bar{\theta}_{r}(t)$ is actually the solution of the PDE in~Equation.~\eqref{PDE}. 
Finally, they showed the distance between two trajectories $\Bar{\theta}_{r}(t)$ and $\theta_{r}(t)$ can be controlled.
In other words, the above PDE~Equation.~\eqref{PDE} describes the evolution of each particle $(\theta_{r})$ in the force field created by the densities of all the other particles.

\section{Beyond the Kernel Regime} \label{AppendixC}
Despite nice theoretical descriptions on training dynamics of gradient descent in loss functions,~\cite{arora2019exact, lee2018deep, chizat1812note} empirically found significant performance gaps between NTK and actual training. 
These gaps have been theoretically studied in~\cite{wei2019regularization, allen2019can, ghorbani2020neural, yehudai2019power} which established that NTK has provably higher generalization error than training the neural net for specific data distributions and architectures.
Here, we introduce two works~\cite{ghorbani2020neural,wei2019regularization}. \\

\begin{enumerate}
    \item \cite{ghorbani2020neural} gives an example of a highly stylized spiked feature model. 
    Consider the case in which $\mathbf{x}=\mathbf{Uz_{1}}+\mathbf{U^{\perp}z_{2}}$, where $\mathbf{U}\in\mathbb{R}^{d\times d_{0}}$, $\mathbf{U}^{\perp}\in\mathbb{R}^{d\times (d-d_{0})}$, and $[\mathbf{U}\mid\mathbf{U}^{\perp}]\in\mathbb{R}^{d\times d}$ is an orthogonal matrix. 
    Set the signal dimension $d_{0}=\lfloor{d^{\eta}}\rfloor$ with $\eta\in(0,1)$ so that $d_{0} \ll d$.
    Here, we call $\mathbf{z}_{1}$ a signal feature and $\mathbf{z}_{2}$ a junk feature
    with the variance $\textbf{Cov}(\mathbf{z}_{1})=\text{SNR}_{f}\mathcal{I}_{d_{0}}$ and $\textbf{Cov}(\mathbf{z}_{2})=\mathcal{I}_{d-d_{0}}$ where  $\text{SNR}_{f}=d^{\kappa}$ for $0\leq \kappa<\infty$.
    Under this setting, the noisy response has the form $\mathbf{y} = \psi(\mathbf{Uz_{1}})+\varepsilon$ with $\varepsilon$ some random noises.
    Note that the response only depends on the signal feature. 
    They derived and compared the number of parameters needed for approximating degree $\ell$-polynomials in $\mathbf{z_{1}}$ in Neural Network (NN), Random Features (RF), and Neural Tangent (NT) models under different levels of $\text{SNR}_{f}$:

    \begin{table}[htbp]
    \centering
    \begin{adjustbox}{width=0.7\textwidth}
    \begin{tabular}{|c|c|c|c||c|}
    \hline
     $\mathcal{F}$ & \quad NN \quad & \quad RF \quad & NT  & $\inf_{f\in\mathcal{F}}\|f^{\star}-f\|_{L^{2}}^{2}$ \\ \hline
     $\text{SNR}_{f}=1$ &  $d^{\eta \ell}$ & $d^{\ell}$  & $d^{\ell}$  & NN $>$ RF = NT \\ \hline
     $\text{SNR}_{f}>1$ &  $d^{\eta \ell}$ & $d^{\eta \ell}$ & $d^{\eta (\ell-1) + 1}$ &  NN $\sim$ RF $>$ NT \\ \hline
    \end{tabular}
    \end{adjustbox}
    \end{table}
    Here, NN is a collection of shallow network functions $f_{NN}(\mathbf{x}, a,\mathbf{W})=\sum_{r=1}^{M}a_{r}\sigma(\langle w_{r}, \mathbf{x} \rangle)$ for $a_{r}\in\mathbb{R}$ and $w_{r}\in\mathbb{R}^{d}$, 
    and RF (resp. NT) are a collection of linear functions $f_{RF}(\mathbf{x}, a) = \sum_{r=1}^{M}a_{r}\sigma(\langle w_{r}^{0}, \mathbf{x} \rangle)$ for $a_{r}\in\mathbb{R}$. (resp. 
    $f_{NT}(\mathbf{x},\mathbf{W})=\sum_{r=1}^{M} \langle s_{r}, \mathbf{x} \rangle \sigma^{\prime}(\langle w_{r}^{(0)}, \mathbf{x} \rangle)$ for $s_{r}\in\mathbb{R}^{d}$.)
    
    It is interesting to observe that the approximation power of NN models is independent of $\text{SNR}_{f}$, whereas the other two models are affected by it.
    In their numerical simulation (Figure $2$), they also showed larger $\text{SNR}_{f}$ induces the larger approximation power of $\{\text{RF, NT}\}$ as manifested in the above table.
    This is also consistent with the spectral bias of NT models in that the model learns the low-frequency components of functions faster than the high-frequency counterparts.

    \item \cite{wei2019regularization} gives an interesting example of where NTK or any kernel methods are statistically limited, whereas regularized neural networks have better sample complexity. 
    Consider the setting where we have feature space $\mathbf{x}\in\mathbb{R}^{d}$ with $\mathbf{x}_{i}\sim\{\pm1\}$ and the function $\mathbf{f}(\mathbf{x})=\mathbf{x}_{1}\mathbf{x}_{2}$ which is only the product of first two arguments of $\mathbf{x}$.
    We want to learn the $\mathbf{f}(\mathbf{x})$ through functions in NTK-induced RKHS $\mathbf{f}^{\text{NTK}}(a;\mathbf{x})=\sum_{r=1}^{M}a_{r}\mathbf{K}^{\infty}(\mathbf{x}_{r},\mathbf{x})$ and two-layer neural networks with ReLU activation $\mathbf{f}^{\text{NN}}(\Theta;\mathbf{x})=\sum_{r=1}^{M}a_{r}\sigma(w_{r}^{\top}\mathbf{x})$, respectively.
    The classifier $\mathbf{f}^{\text{NTK}}(a;\mathbf{x})$ is attained by minimizing squared loss and $\mathbf{f}^{\text{NN}}(\Theta;\mathbf{x})$ is estimated through $\ell_{2}$-regularized logistic loss.
    Under this setting, Theorem $2.1$ in~\cite{wei2019regularization} is read as:
    \begin{center}
        ``Functions $\mathbf{f}^{\text{NTK}}(a;\mathbf{x})$ require $n=\Omega(d^{2})$ samples to learn the problem with error $\ll\Omega(1)$. In contrast, regularized NN functions $\mathbf{f}^{\text{NN}}(\Theta;\mathbf{x})$ only need $n=\mathcal{O}(d)$ samples.''
    \end{center}
    The result implies that there is a $\Omega(d)$ sample-complexity gap between the regularized neural net and kernel prediction function.
    A main intuition behind this gap is that the regularization allows neural networks to adaptively find the model parameters $(a_{r},w_{r})$ so that the best estimator only chooses $4$ neurons in the hidden layer. {\footnote{Using the $|t|:=\sigma(t) + \sigma(-t)$ for $t\in\mathbb{R}$, we can easily see the function $\mathbf{y}=\mathbf{x_{1}}\mathbf{x_{2}}$ can be re-written as $\mathbf{y} = \frac{1}{2}\big[ \sigma(\mathbf{x_{1}}+\mathbf{x_{2}}) + \sigma(-\mathbf{x_{1}}-\mathbf{x_{2}}) - \sigma(\mathbf{x_{1}}-\mathbf{x_{2}}) - \sigma(\mathbf{x_{2}}-\mathbf{x_{1}})\big]$.}}\
    (Note that the relation between $\ell_{2}$-regularized two-layer neural network and $\ell_{1}$-SVM over ReLU features is known in literature~\cite{neyshabur2014search}.)
    However, using the NTK, we do a dense combination of existing features as the network output will always involve $\sigma(w^{\top}\mathbf{x})$ where $w$ is a random vector so it includes all the components of $\mathbf{x}$. \\
\end{enumerate}

\noindent \textit{\textbf{{Beyond the kernel regime.} }}
The above two works explained the superiority of neural networks over the networks in the NTK regime under some \textit{highly stylized} settings.
There also have been several theoretical attempts~\citep{bai2019beyond,allen2019learning} to explain how the networks estimated through gradient-based methods generalize well to the unseen data but critically do not rely on the linearization of network dynamics. 
The paper~\cite{bai2019beyond} studied the optimization and generalization of shallow networks with smooth activation function $\sigma(\cdot)$ via relating the network dynamics $f_{\mathbf{W}}(\mathbf{x})$ with higher-order approximations. The key idea is to find the subset of $\mathbf{W}:=\{(a_{r}, w_{r})\}_{r=1}^{M}$ such that the dominating term in Taylor expansion~Equation.~\eqref{NTK_Approx} is not the linear term (i.e., $\langle \nabla f_{\mathbf{W}_{0}}(\mathbf{x}), \mathbf{W}-\mathbf{W}_{0} \rangle$) but a quadratic term (i.e., $\| \mathbf{W}-\mathbf{W}_{0} \|_{\text{F}}^{2}$).
The authors find such $\mathbf{W}$ by running SGD on the tailored-regularized loss:
\begin{align}\label{Landscape}
    \mathbb{E}_{\mathbf{\Sigma},(\mathbf{x},\mathbf{y})\sim\mathcal{D}}\bigg[
    \ell(\mathbf{y}, f_{\mathbf{W}_{0}+\mathbf{\Sigma W}_{r}}(\mathbf{x}))\bigg] + \lambda \|\mathbf{W}\|_{2,4}^{8}.
\end{align}

\noindent
The expectation in loss is taken over the diagonal matrix $\mathbf{\Sigma}$ with $\Sigma_{rr}\sim\text{Unif}\{-1,1\}$ over $r\in[M]$ and it has the effects of dampening out the linear term.
The regularizer $\|\cdot\|_{2,4}^{8}$ controls the distance of moving weights to be $\mathcal{O}(M^{-1/4})$.
The authors showed the landscape of optimization problem~Equation.~\eqref{Landscape} has a very nice property that every second-order stationary point of the loss is nearly global optimum, despite the non-convexity of~Equation.~\eqref{Landscape}.
These points can be easily found by the SGD algorithm.
Given that $M$ is sufficiently large enough, i.e., $M\gtrsim n^{4}$, they further provided the comparisons of sample complexities between quadratic and linear models under three different scenarios.
The ground truth functions estimated in the three scenarios are 
(1) $f^{\star}(\mathbf{x})=\alpha(\beta^{\top}\mathbf{x})^{p}$ for $\alpha\in\mathbb{R}$,  $\beta\in\mathbb{R}^{d}$, and $p=1 \text{ or even}$. 
(2) $f^{\star}(\mathbf{x})=\mathbf{x}_{1}\mathbf{x}_{2}$, and 
(3) $f^{\star}(\mathbf{x})=\langle \Theta^{\star}, \mathbf{x}\mathbf{x}^{\top}\rangle$ where $\Theta^{\star}\in\mathbb{R}^{d \times d}$ is a rank-$r$ matrix.
We summarize the results in the below table.
\begin{table}[htbp]
\centering
\begin{adjustbox}{width=0.9\textwidth}
\begin{tabular}{|c|c|c|c|}
\hline
 & $\mathcal{X}$ & \textbf{Quadratic Model} & \textbf{Linear Model} \\ \hline
 $f^{\star}(\mathbf{x}) = \alpha(\beta^{\top}\mathbf{x})^{p}$ 
 & Sufficiently Isotropic & $n\geq\Tilde{\mathcal{O}}\bigg(\frac{p^{3}\alpha^{2}\|\beta\|_{2}^{2p}}{d\varepsilon^{2}}\bigg)$   
 & $n\geq\Tilde{\mathcal{O}}\bigg(\frac{p^{2}\alpha^{2}\|\beta\|_{2}^{2p}}{\varepsilon^{2}}\bigg)$  \\ \hline
 $f^{\star}(\mathbf{x}) = \mathbf{x}_{1}\mathbf{x}_{2}$ 
 & $\{-1,+1\}^{d}$ & $n\geq\Tilde{\mathcal{O}}\bigg(\frac{d}{\varepsilon^{2}}\bigg)$ 
 & $n\geq\Omega(d^{2})$ 
 \\ \hline
 $f^{\star}(\mathbf{x}) = \langle \Theta^{\star}, \mathbf{x}\mathbf{x}^{\top}\rangle$ 
 & $\text{Unif}(\mathbf{S}^{d-1}(\sqrt{d})$ & $n\geq\Tilde{\mathcal{O}}\bigg(\frac{dr^{2}}{\varepsilon^{2}}\bigg)$ 
 & $n\geq\Tilde{\mathcal{O}}\bigg(\frac{d^{2}r^{2}}{\varepsilon^{2}}\bigg)$ \\
\hline
\end{tabular}
\end{adjustbox}
\end{table}

In all three cases, it is shown that the lower bound of sample complexity of the quadratic model is smaller than that of the linear NTK model by the factor of $\Tilde{\mathcal{O}}(d)$.
This can mean that the expressivity of the higher-order approximation of neural dynamics is richer than that of the linear counterpart. 

\section{Unifying View of NTK and Mean-Field Regime} \label{AppendixD}
There have been attempts to give a unifying view of these two regimes. 
The first attempt we are aware of is~\cite{chen2020generalized}. 
The motivation of this paper is summarized in the above table, along with the pros and cons of NTK and Mean-Field regimes. 
As mentioned in Table~\ref{Table1}, the caveat of MF analysis lies in its difficulties in obtaining the generalization bound, yet the regime allows the parameters to travel far away from their initializations. 
The key idea of the paper is to bring in the NTK which requires $\|\theta^{(k)}-\theta^{(0)}\|_{\text{2}}^{2}$ to be small.
Instead, they worked on the probability measure space $\mathcal{P}(\mathbb{R}^{d})$ minimizing the following energy functional, where $f(\mathbf{x};\rho):=\alpha \int \sigma_{\star}(\mathbf{x};\mathbf{\theta})\rho(d\mathbf{\theta})$ is the integral representation of neural dynamics in~Equation.~\eqref{Shallow}:
\begin{align}\label{Functional}
    \rho^{\star} := \argmin_{\rho\in\mathcal{P}(\mathbb{R}^{d})} \bigg\{\frac{1}{n}\sum_{i=1}^{n}\big(
    \mathbf{y}_{i}-f(\mathbf{x}_{i};\rho)\big)^{2}+\lambda \mathcal{D}_{\textbf{KL}}(\rho\parallel\rho_{0})\bigg\},
\end{align}
with a hyper-parameter $\lambda>0$.
Here, $\rho^{\star}\in\mathcal{P}(\mathbb{R}^{d})$ is a global minimizer of Equation.~\eqref{Functional}.
Note that they penalize the KL-divergence between $\rho$ and $\rho_{0}$ instead the distance between $\theta^{(k)}$ and $\theta^{(0)}$.
The training of the objective functional~Equation.~\eqref{Functional} is often implemented through the Wasserstein gradient flow~\cite{mei2018mean}. 
Note that the noisy gradient descent (NGD) corresponds to a discretized version of the Wasserstein gradient flow in parameter space, and it has been extensively studied in~\cite{mei2018mean,mei2019mean,chizat2018global} that the NGD algorithm approximates the joint measure $\rho^{\star}$ which is the solution of~Equation.~\eqref{PDE} with an additional diffusion term. 
\\ \\
Under this setting,~\cite{chen2020generalized} showed the generalization bound for the binary classification problem with $0-1$ loss. (i.e., $\ell^{0-1}(y,y^{\prime}):=\mathbbm{1}(yy^{\prime}<0)$.)
Given the target function has the form $\mathbf{y}:=f(\mathbf{x};\rho_{\text{true}})$, the generalization bound is given as:  
\begin{align}
    \mathbb{E}_{\mathcal{D}}\big[\ell^{0-1}\big(f(\rho^{\star},\mathbf{x}),\mathbf{y}\big)\big]
    \leq \underbrace{\Tilde{\mathcal{O}}\bigg( \sqrt{\frac{\mathcal{D}_{\mathcal{X}^{2}}(\rho_{\text{true}}||\rho_{0}\big)}{n}} \bigg)}_{\text{NTK} \big( \alpha=\mathcal{O}(\sqrt{M})\big)},
    \quad 
    \underbrace{\Tilde{\mathcal{O}}\bigg( \sqrt{\frac{\mathcal{D}_{\text{KL}}(\rho_{\text{true}}||\rho_{0}\big)}{n}} \bigg)}_{\text{MF} \big( \alpha=\mathcal{O}(1)\big)}.
\end{align}
This implies that when $\alpha$ in~Equation. 8 (in main text) is large, i.e., $\alpha=\mathcal{O}(\sqrt{M})$ (resp. $\alpha=\mathcal{O}(1)$), two-layer neural networks with infinite width trained by noisy gradient descent (NGD) can learn the function class $\mathcal{F}_{\mathcal{X}^{2}}$ (resp. the function class $\mathcal{F}_{\textbf{KL}}$).
The two function classes are defined as follows:
\begin{align*}
    \mathcal{F}_{\mathcal{X}^{2}}:=\bigg\{ \int \sigma_{\star}(\mathbf{x};\mathbf{\theta})\rho(d\mathbf{\theta})
    : \mathcal{D}_{\mathcal{X}^{2}}(\rho \parallel \rho_{0})<\infty\bigg\},
    \quad 
    \mathcal{F}_{\textbf{KL}}:=\bigg\{ \int \sigma_{\star}(\mathbf{x};\mathbf{\theta})\rho(d\mathbf{\theta})
    : \mathcal{D}_{\textbf{KL}}(\rho \parallel \rho_{0})<\infty\bigg\},
\end{align*}
where $\mathcal{D}_{\mathcal{X}^{2}}(\rho\parallel \rho_{0})$ and $\mathcal{D}_{\textbf{KL}}(\rho\parallel \rho_{0})$ denote the $\mathcal{X}^{2}$ and $\textbf{KL}$-divergences between $\rho$ and $\rho_{0}$, respectively.
As KL-divergence is smaller than $\mathcal{X}^{2}$ distance, it can be checked $\mathcal{F}_{\mathcal{X}^{2}} \subsetneq \mathcal{F}_{\textbf{KL}}$ (a notation $\subsetneq$ denotes a strict subset).
This implies that the two-layer neural networks learned through NGD in mean-field scaling can potentially learn a larger class of functions than networks in NTK scaling can learn. 
\begin{table}[!t]
\centering
\begin{tabular}{|cc|c|c|c|}
\hline
\multicolumn{1}{|c|}{}   & \textbf{Definition}  & \textbf{NTK} & \textbf{Mean Field (L=1)} & \textbf{Maximal Update ($\mu P$)} \\ \hline
\multicolumn{1}{|c|}{\textbf{$a_{\ell}$}}   &  $W_{\ell}=M^{-a_{\ell}}w^{\ell}$           & $\begin{cases}
    0   \qquad \ell=1 \\
    1/2 \quad \ell>1
\end{cases}$    &       
$\begin{cases}
    0   \quad \ell=1 \\
    1   \quad \ell=2
\end{cases}$           &       
$\begin{cases}
    -1/2   \qquad \ell=1 \\
    0   \quad \qquad  2 \leq \ell \leq L \\
    1/2 \quad \quad \ell = L+1 \\   
\end{cases}$  
\\ \hline
\multicolumn{1}{|c|}{\textbf{$b_{\ell}$}}   &  $w_{\alpha \beta}^{\ell}\sim \mathcal{N}(0,M^{-2b_{\ell}})$           &   0  &       0      &    1/2        \\ \hline
\multicolumn{1}{|c|}{\textbf{$c_{\ell}$}}   &   $LR=\eta M^{-c}$    &  0   &  -1   & 0            \\ 
\hline
\end{tabular}
\end{table}
\\ \\
Another work~\cite{yang2020feature} provides a unifying framework encompassing the parametrizations that induce the neural dynamics in NTK and MF regimes. 
Under the heavy overparameterized setting (i.e., the width of the network is very large), the main differences between these two regimes come from the different scales on {initialized weight parameters} and {learning rates} in GD or SGD. 
In the paper, the proposed \textit{abc-parameterization} provides a guideline on how to choose the scales on these parameters.
Specifically, recall the definition of a fully connected network with $L+1$ layers in~Equation. 1 (in main text).
In this context, the width of the network remains consistent at $M$ across all layers.
The parametrization is given by a set of numbers $\{a_{\ell},b_{\ell}\}_{\ell} \cup \{c\}$.
The pair $(a_{\ell},b_{\ell})$ controls the scale of initialized weight parameters over the layers, and $c$ is the parameter relevant to the learning rate of SGD. 

The parameterization is given as: 
\begin{enumerate}
    \item Parameterize $W_{\ell} = M^{-a_{\ell}}w^{\ell}$ where $w^{\ell}$ is trained instead of $W_{\ell}$ for all $\ell\in[L+1]$.
    \item Initialize each $w_{\alpha \beta}^{\ell}\sim\mathcal{N}(0,M^{-2b_{\ell}})$.
    \item SGD learning rate is $\eta M^{-c}$ for some width-independent $\eta>0$.
\end{enumerate}
Borrowing the paper's notation, let us denote $\mathbf{x}_{t}^{\ell}(\mathbf{\xi})$ as the output of the $\ell$-th hidden layer of fully-connected network~Equation. 1 (in main text) (after the activation) with the input $\mathbf{\xi}\in\mathbb{R}^{d}$ for the $t$-th SGD updates. 
The paper defines the difference vector $\Delta\mathbf{x}_{t}^{\ell}(\mathbf{\xi}):=\mathbf{x}_{t}^{\ell}(\mathbf{\xi})-\mathbf{x}_{0}^{\ell}(\mathbf{\xi})$ for all $\ell\in[L+1]$ with $r$ notation:
\begin{align}
    r:=\min(b_{L+1}, a_{L+1}+c)+a_{L+1}+c+\min_{\ell=1,\dots,L}[2a_{\ell}-\mathbbm{1}(\ell \neq 1)].
\end{align}

Under this setting, they consider the regimes of parameterizations $\{a_{\ell},b_{\ell}\}_{\ell} \cup \{c\}$ where every coordinate of $\mathbf{x}_{t}^{\ell}(\mathbf{\xi})$ doesn't blow-up for every $\ell\in[L+1]$.
Specifically, the regimes for which $r\geq 0$ when $\Delta\mathbf{x}_{t}^{\ell}(\mathbf{\xi}):=\Theta(M^{-r})$ are considered.
This is in conjunction with the additional linear constraints on $\{a_{\ell},b_{\ell}\}_{\ell} \cup \{c\}$ where the network outputs at initialization and during training don't blow-up.
(See Theorem 3.2 in~\cite{yang2020feature} for the detail.)
The high-dimensional polyhedron formed through these constraints is referred to as {a stable abc-parameterization} region in the sense that training through SGD should be stable avoiding the blow-up of network dynamics.
They further showed the stable region is divided into two parts where the dynamics change ({non-trivial}) and doesn't change ({trivial}) in its infinite width limit, i.e., $M\rightarrow\infty$ during training. 
The {non-trivial stable abc-parameterization} is possible on the union of the following two faces of the polyhedron:
\begin{align} \label{constr_poly}
    a_{L+1}+b_{L+1}+r=1 \quad \text{or} \quad 2a_{L+1}+c=1.
\end{align}
In this region, the infinite width limit either (1) allows the embedding of the last hidden layer $\mathbf{x}^{L}(\mathbf{\xi})$ to evolve non-trivially with $r=0$ (MF regime) or (2) is described by kernel gradient descent with $r>0$ (NTK regime), but not both. 
\\ \\
The corresponding settings of $(a_{\ell},b_{\ell})\cup \{c\}$ for NTK and MF regimes are summarized in the above table.
They only focus on the scaling with $M$ and ignore dependence on the intrinsic dimension $d$.
The setting for NTK matches with the Theorem $5$ in~\cite{du2019gradient}, where the entries of $W_{\ell}$ is drawn from $\mathcal{N}(0,1/M)$ with width-independent $\eta>0$. 
The reason why we are seeing $a_{\ell}=0$ for $\ell=1$ in the NTK regime is to ensure that the output of every hidden layer in the network is of constant order so that parameterization is in a {stable} regime. 
For the MF regime,~\cite{mei2018mean,mei2019mean} gives us the width-independent parameterizations for weight initializations and learning rate in {shallow} neural network.
Note that the $\frac{1}{M}$-scale in the output weight layer $(\ell=2)$ is canceled with the $M$-scale of learning rate in SGD.
\\ \\
Although not mentioned in the paper, we conjecture the reason for these manipulations is to put the parameterizations in the MF regime to satisfy the constraints~Equation.~\eqref{constr_poly}. 
The $\{a_{\ell},b_{\ell}\}_{\ell} \cup \{c\}$ pair for NTK satisfies $r=\frac{1}{2}$, meaning the dynamic is described through kernel gradient descent, whereas the pair for MF satisfies $r=0$, allowing the feature learning.
The authors of this paper showed the feature learning dynamics in the MF regime can be extended to a multi-layer setting, where they call the corresponding parameterizations as \textbf{M}aximal \textbf{U}pdate parameterization ($\mu P$).
This parametrization is ``\textbf{Maximal}'' in the sense that feature learning is enabled throughout the entire layers in the network in the $\infty$-width limit.
As mentioned in the introduction, feature learning is the essence of modern deep learning making techniques such as transfer learning or fine-tuning in the Large Language Model (LLM) possible, and  
it is remarkable in the sense that the exact neural dynamics are mathematically tractable under this setting. 

\end{appendices}

\bibliographystyle{plainnat}
\bibliography{main}

\end{document}